\documentclass[preprint,12pt,authoryear]{elsarticle}

\usepackage[noend]{algpseudocode}
\usepackage{algorithmicx,algorithm}

\algnewcommand{\algorithmicforeach}{\textbf{for each}}
\algdef{SE}[FOR]{ForEach}{EndForEach}[1]
  {\algorithmicforeach\ #1\ \algorithmicdo}
  {\algorithmicend\ \algorithmicforeach}

\usepackage[final]{changes}

\usepackage{amssymb}
\usepackage{makecell}
\usepackage{booktabs}
\usepackage{graphicx}
\usepackage{multirow}
\usepackage{float}
\usepackage{subfig}
\usepackage{color}
\usepackage{amsmath,bm}

\usepackage{lineno}

\usepackage{hyperref}

\usepackage{color, soul}
\setulcolor{red}
\setstcolor{blue}
\sethlcolor{yellow}

\begin{document}

\begin{frontmatter}



\title{Efficient Feature Matching for UAV Images based on Compact GPU Data Scheduling}


\author[label1,label2,label3]{San Jiang}
\author[label3]{Kan You}
\author[label4]{Ruqin Zhou}
\author[label1,label2]{Xing Zhang}
\author[label5]{Zhijun Wang}
\author[label1,label2]{Qingquan Li\corref{cor1}}

\affiliation[label1]{organization={Guangdong Key Laboratory of Urban Informatics, Shenzhen University},
            city={Guangdong Shenzhen},
            postcode={518060}, 
            country={China}}

\affiliation[label2]{organization={MNR Key Laboratory for Geo-Environmental Monitoring of Great Bay Area, Shenzhen University},
            city={Shenzhen},
            postcode={518060}, 
            country={China}}
            
\affiliation[label3]{organization={Shenzhen Key Laboratory of Spatial Smart Sensing and Services},
            city={Shenzhen},
            postcode={518060}, 
            country={China}}

\affiliation[label4]{organization={School of Surveying and Mapping, Information Engineering University},
            city={Zhengzhou},
            postcode={450001}, 
            country={China}}

\affiliation[label5]{organization={Guangdong Laboratory of Artificial Intelligence and Digital Economy (Shenzhen)},
            city={Shenzhen},
            postcode={518060}, 
            country={China}}

\cortext[cor1]{Corresponding Author: liqq@szu.edu.cn}

\begin{abstract}
Feature matching dominates the time costs in structure from motion (SfM). The primary contribution of this study is a GPU data schedule algorithm for efficient feature matching of Unmanned aerial vehicle (UAV) images. The core idea is to divide the whole dataset into blocks based on matrix band reduction (MBR) and achieve efficient feature matching via GPU-accelerated cascade hashing. First, match pairs are selected by using an image retrieval technique, which converts images into global descriptors and searches high-dimension nearest neighbors with graph indexing. Second, compact image blocks are iteratively generated from a MBR-based data schedule strategy, which exploits image connections to generate image blocks and increase the usage of GPU computing power. Third, guided by the generated image blocks, feature matching is executed sequentially within the framework of GPU-accelerated cascade hashing, and initial candidate matches are refined by combining a local geometric constraint and RANSAC-based global verification. For further performance improvement, these two steps are designed to execute in parallel in GPU and CPU. Finally, the performance of the proposed solution is evaluated by using large-scale UAV datasets. The results demonstrate that it increases the efficiency of feature matching with speedup ratios ranging from 77.0 to 100.0 compared with KD-Tree based matching methods due to its high usage of GPU computing power. Besides, it achieves comparable accuracy in both relative and absolute bundle adjustment (BA). The proposed algorithm is an efficient solution for feature matching of large-scale UAV images.
\end{abstract}




\end{frontmatter}


\section{Introduction}
\label{sec1}
Unmanned aerial vehicle (UAV) has been increasingly becoming a critical remote sensing platform in photogrammetry and computer vision \citep{jiang2021unmanned, yao2019unmanned}, and it has been widely used in a range of applications, e.g., indoor inspection of transmission lines \citep{jiang2017uav}, procedural 3D modeling of urban facilities \citep{wang2023oblique}, and precision management of agricultural plants \citep{zheng2021growing}. The precise calculation of camera poses is an essential preliminary step, which has been usually implemented through structure from motion (SfM)  \citep{jiang2020effcient}. In the workflow of SfM, the time costs are dominated by feature matching due to the quadratic combination complexity of match pairs and nearest neighbor search (NNS) of high-dimension feature descriptors. Therefore, fast feature matching is required for recent SfM systems.

\begin{figure*}[!tp]
    \centering
    \includegraphics[width=1.0\linewidth]{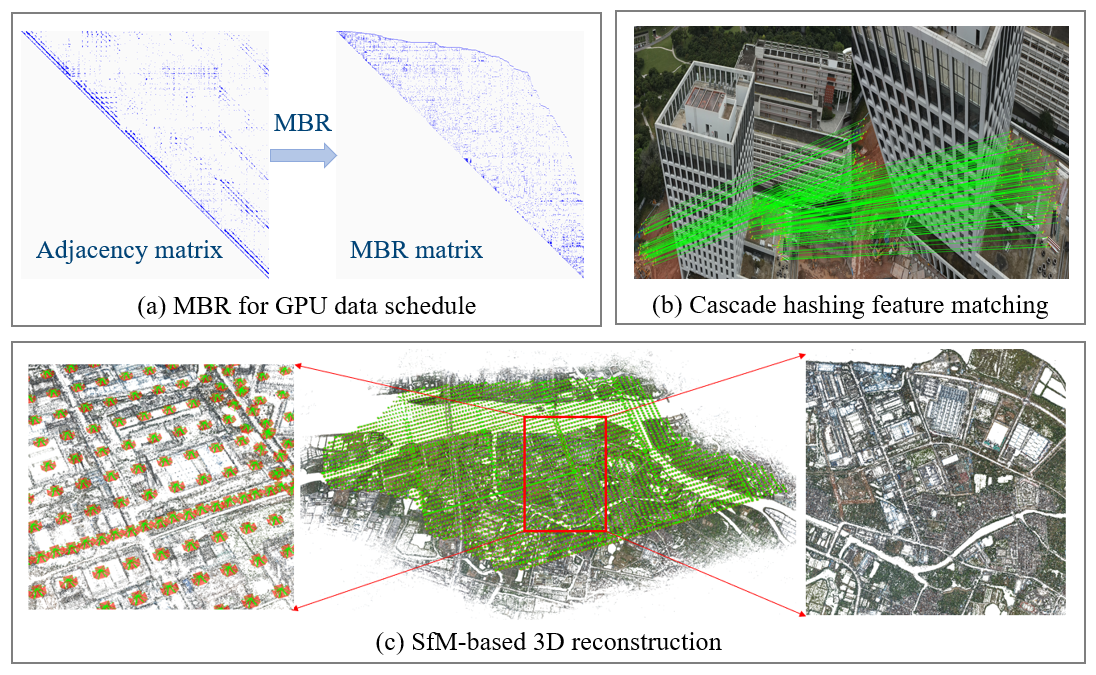}
    \caption{GPU data schedule algorithm for feature matching and SfM reconstruction.}
    \label{fig0}
 \end{figure*}

In the literature, extensive work has been reported to address these issues. To decrease the combination complexity of match pairs, auxiliary data and visual similarity are the widely used clues to select overlapped image pairs and avoid exhaustive feature matching \citep{hartmann2016recent}, e.g., the onboard POS (Positioning and Orientation System) data from UAV platforms \citep{jiang2020efficient,remondino2015oblique} and the geometric priors from data acquisitions \citep{jiang20243d,schonberger2014structure}. In addition, to cope with oblique data acquisition and geometry-aware trajectory planning \citep{li2023optimized}, visual similarity has also been extensively exploited in modern SfM systems, which usually use pre-trained visual words to encode images into fix-length vectors and convert the problem of selecting overlapped image pairs into the one of nearest neighbor search among the vectors, such as BoW (Bag-of-Words) \citep{nister2006scalable,havlena2014vocmatch,jiang2022leveraging} and VLAD (Vector of Locally Aggregated Descriptors) \citep{hou2023learning,jiang2023efficient}. By using these techniques, the time complexity of feature matching is significantly decreased, which is transited from a quadratic relationship to a linear one with respect to the number of involved UAV images.

Match pair selection reduces the combination complexity of images. However, high time costs are still required in the exhaustive nearest neighbor search among two sets of high-dimension feature descriptors \citep{hartmann2016recent}, especially for the widely used KD-Tree based SIFT feature matching \citep{lowe2004distinctive}. \added{Nowadays, deep learning based methods have also been reported to achieve reliable feature matching, such as SuperGlue \citep{sarlin2020superglue} and LoFTR \citep{sun2021loftr}. Although they avoid nearest neighbor searching, the localization precision still the main drawback for their usage in SfM.} To accelerate feature matching, existing methods can be divided into two groups, i.e., redesign of feature descriptors and acceleration of feature matching. For the former, feature descriptors are designed to decrease the computational burden, such as the ORB \citep{rublee2011orb} with binary descriptors. However, as verified in recent work, SIFT-based feature matching is still the most robust algorithm in practical engineering applications \citep{ji2023evaluation,jiang2021learned}. For the latter, the hardware acceleration technique has been exploited to speed up feature matching. The classical examples include the SIFTGPU \citep{wu2011siftgpu} and popsift \citep{griwodz2018popsift} algorithms that exploit GPU (Graphic Processing Unit) for feature matching. In contrast to the KD-Tree based nearest neighbor search, hashing-based methods have also gained lots of attention in feature matching, whose core idea is to convert the real-value descriptors to discrete binary codes via hashing functions and map similar descriptors of query and database images into the same cell \citep{cheng2014fast}. To achieve higher speedup ratios, GPU acceleration and data schedule have been integrated into hashing-based matching pipelines in the recent work \citep{xu2017gpu,zhang2023efficient}, which use hardware acceleration for calculating hash codes and similarity scores and exploit data schedule to relief input/output (IO) burdens.

Compared with classical SIFT matching algorithms, hashing-based methods can achieve promising speedup ratios for feature matching of UAV images. However, some issues still exist in the proposed solutions. First, the performance of existing data schedule depends on dense image connections. Pair-wise feature matching is first executed for all images to create an image adjacency matrix, and the group-block based data schedule strategy is used to divide the whole dataset into groups and blocks that are fed into GPU sequentially for feature matching \mbox{\citep{xu2017gpu}}. However, for oblique photogrammetry, image connections are usually very sparse, and it degenerates the overall performance of GPU-based hashing matching since a majority of blocks have few image pairs. Second, existing data schedule strategies emphasize more on the IO burdens, however, ignoring the parallel computing power of modern GPU. For the data schedule used, image loading and free sequences are first generated based on the image connection structure, and feature matching is executed sequentially according to loading and free sequence \mbox{\citep{zhang2023efficient}}. In other words, only one image is matched against other images in GPU memory. Third, outliers are prone to retain in feature matching due to the usage of binary hashing codes for calculating descriptor similarity scores. It would cause high time cost in classical RANSAC (Random Sample Consensus) based outlier removal \citep{lu2016geometrical}. 

To address above-mentioned issues, this study proposes a matrix band reduction-based GPU data schedule for fast feature matching of UAV images, as shown in Figure \ref{fig0}. Our main contributions are summarized as follows: (1) we propose a matrix band reduction (MBR) based data schedule strategy to divide the sparsely connected view graph into compact blocks, which can adapt well to the connection structure of images and the memory volume of GPUs; (2) we design a cascade hashing based feature matching workflow that integrates the spatial angular order (SAO) based local constraint and RANSAC-based global verification for outlier removal. Combined with MBR-based data schedule, the proposed image matching solution can increase the usage of GPU for feature matching and CPU for outlier removal; (3) we verify the performance of the proposed feature matching solution using large-scale datasets and compare it with state-of-the-art software packages.

This paper is organized as follows. Section \ref{sec3} presents the workflow of the GPU data schedule algorithm for efficient feature matching of UAV images. Section \ref{sec4} conducts the performance evaluation and comparison with other methods. Finally, Section \ref{sec6} presents the conclusions.

\begin{figure*}[!tp]
    \centering
    \includegraphics[width=1.0\linewidth]{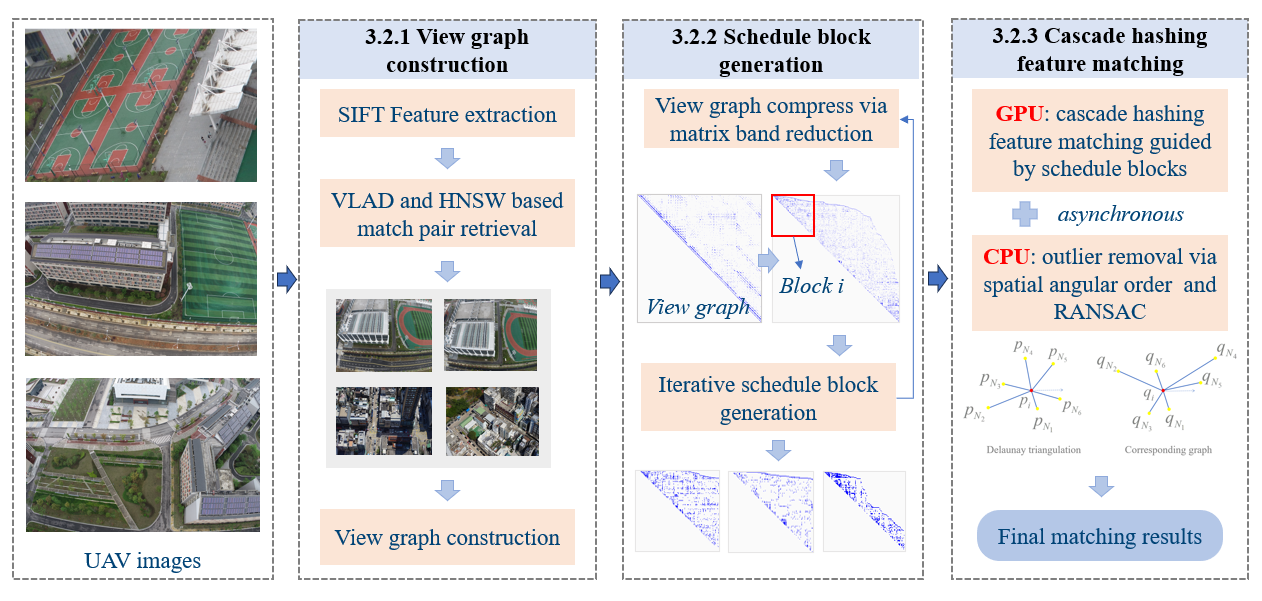}
    \caption{The workflow of the proposed algorithm.}
    \label{fig1}
 \end{figure*}

\section{Methodology}
\label{sec3}

This study proposes a fast feature matching solution for UAV images via matrix band reduction-based GPU data schedule, as shown in Figure \ref{fig1}. There are three major steps, i.e., view graph construction, schedule block generation, and cascade hashing feature matching. First, SIFT features are extracted from each image and used to calculate VLAD-based image descriptors using a pre-trained codebook. After indexed in the HNSW (Hierarchical Navigable Small World) based graph structure, overlapped image pairs are retrieved and used to construct the view graph. Second, the initial view graph is first compressed through the matrix band reduction algorithm, which permutes the image order and ensures that overlapped match pairs are located near the diagonal region in the view graph. Data schedule blocks are then created from the permuted adjacent matrix. Through the iterative execution of MBR, a list of schedule blocks can be generated. Finally, images in each schedule block are loaded into GPU memory for calculating hashing codes and generating initial matching results. At the same time, outlier removal is simultaneously conducted on the CPU by using spatial angular order-based local geometric constraint and RANSAC-based global geometric constraint. The details are presented in the following sections.

\subsection{View graph construction based on image retrieval}

Match pairs are first selected to create a view graph that would be used for data schedule between hard disk and GPU memory and guide feature matching. Although on-board POS data can be used as useful clues for efficient match pair selection, this method depends on the precision of onboard sensors and cannot adapt to geometry-aware data acquisition, e.g., optimized views photogrammetry. Instead, visual similarity-based image retrieval has gained more attention, and BoW-based image retrieval encodes feature descriptors as vectors by using a pre-trained codebook and casts match pair selection as the problem of searching the nearest neighbors among BoW vectors. It has been implemented in commercial and open-source software packages. However, its efficiency decreases dramatically for large-scale UAV images due to the large number of extracted features and the large-size codebook required for encoding large-scale datasets.

\begin{figure}[!t]
    \centering
    \includegraphics[width=1.0\linewidth]{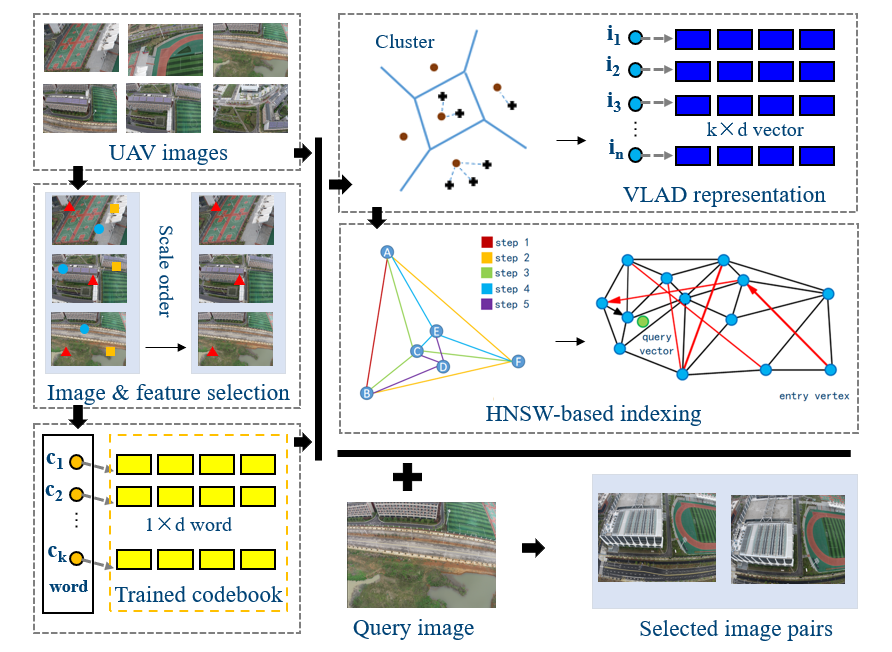}
    \caption{The workflow of the VLAD-HNSW based image retrieval workflow.}
    \label{fig2a}
 \end{figure}

To address this challenge, the study employs a high-efficiency image retrieval framework \citep{jiang2023efficient}, as shown in Figure \ref{fig2a}. The proposed method includes three steps: (1) codebook generation based on image and feature selection; (2) the VLAD  representation for compact image encoding; and (3) the HNSW graph structure for efficient vector indexing. VLAD descriptors utilize a minimal-size codebook to aggregate local feature descriptors into a unified high-dimensional global representation, significantly reducing computational redundancy. The HNSW index architecture accelerates the approximate nearest neighbor search process over the encoded VLAD vectors, enabling rapid similarity matching while maintaining sublinear time complexity. This synergistic combination balances representational compactness with scalable search performance.

According to the VLAD-HNSW based image retrieval, a set of match pairs $P=\left\{p_{ij}\right\}$ are selected and used to create the view graph. Suppose that there are $n$ images, the view graph is constructed as an adjacent matrix $M_{ij},\ i=1,2,\ldots,n;j=1,2,\ldots,n$. For each match pair $p_{ij}$, set corresponding item $M_{ij}=1$; otherwise, set $M_{ij}=0$. In this study, the view graph $M_{ij}$ would be used to generate blocks for data schedule and guide feature matching.

\subsection{Schedule block generation via matrix band reduction}



\begin{figure}[!t]
    \centering
    \includegraphics[width=1.0\linewidth]{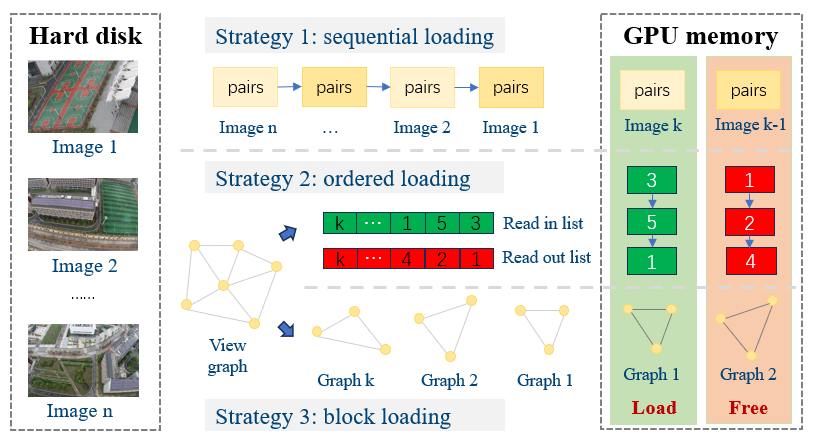}
    \caption{The illustration of existing data schedule strategies between hard disk and GPU memory.}
    \label{fig2b}
 \end{figure}

Data schedule between hard disk and GPU is required because all feature descriptors cannot be loaded into GPU with limited memory at one time. In general, existing strategies for data schedule can be grouped into three categories, as illustrated in Figure \ref{fig2b}. The most direct strategy is to execute feature matching sequentially for all selected match pairs. This strategy, however, causes frequent data load and free in GPU. For the second category, data load and free lists are first generated according to their connections, and feature matching is conducted by the interleaved execution of reading in data in the load list and clearing out data in the free list. This strategy decreases data schedule burdens. However, it does not fully exploit the computing power of GPU since only the newly loaded image is matched with other images in GPU. For the third category, images are divided into blocks that can be loaded into GPU at one time, and feature matching is then executed between loaded images. It has high efficiency in pair-wise feature matching. For sparsely connected images, its performance decreases obviously.

\begin{figure*}[!tp]
    \centering
    \begin{minipage}[t]{1.0\linewidth}
    \centering
        \begin{tabular}{@{\extracolsep{\fill}}c@{}c@{}c@{}@{\extracolsep{\fill}}}
            \includegraphics[width=0.5\linewidth]{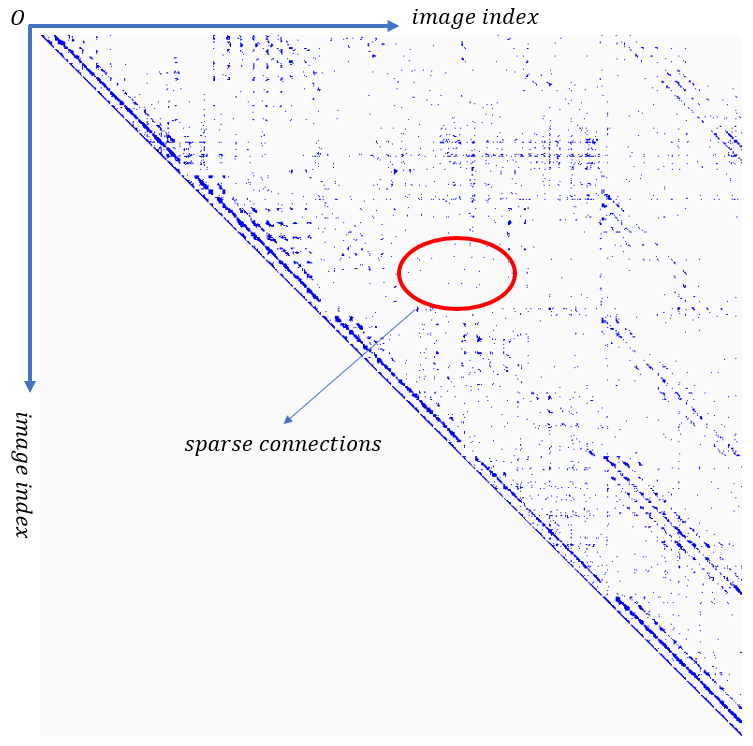} &
            \includegraphics[width=0.5\linewidth]{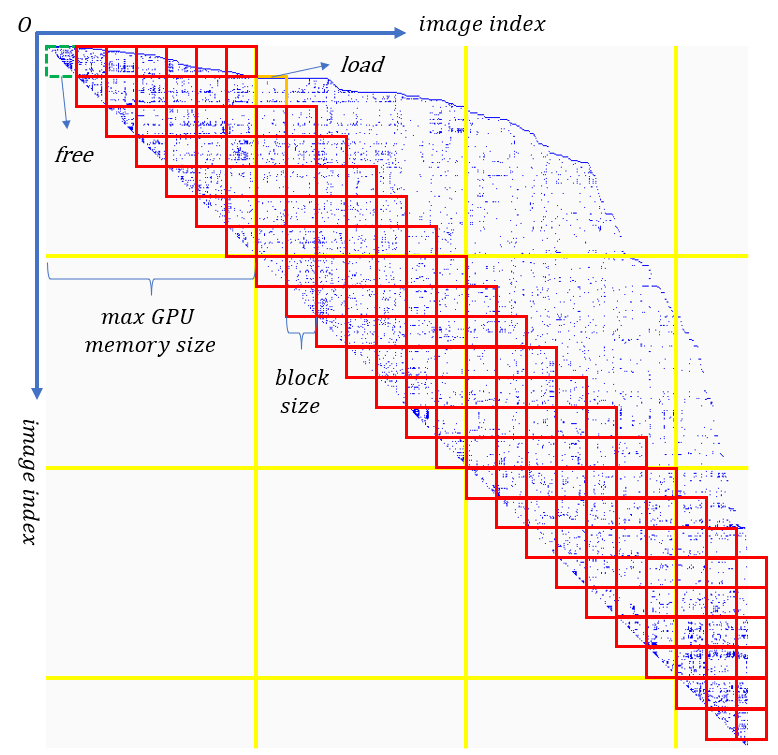}\\
            (a) & (b) \\
        \end{tabular}
    \end{minipage}
   
    \caption{The illustration of schedule block generation based on matrix band reduction: (a) original adjacent matrix; (b) compressed MBR matrix. The blue color indicates connected image pairs while the white color means no connection between corresponding images.}
    \label{fig3}
 \end{figure*}

Inspired by recent work, this study proposes a data schedule strategy based on iterative matrix band reduction. The core idea is to reduce the bandwidth of the original adjacent matrix $M_{ij}$ via a matrix band reduction algorithm and generate schedule blocks from the compressed MBR matrix under the limitation of GPU memory. In the matrix theory, matrix band reduction, often referred to as bandwidth reduction, is a technique used in numerical analysis to optimize the storage and computational efficiency of sparse matrices. The band of a matrix consists of the diagonal and the rows and columns immediately adjacent to it, which contain the non-zero elements of the matrix. Figure \ref{fig3} illustrates the effect of MBR. The original adjacent matrix $M_{ij}$ has sparse image connections that are rendered by blue color and distributed over the whole matrix space, as shown in Figure \ref{fig3}(a). After applying MBR, the image connections are recorded and located near the diagonal as near as possible, as shown in Figure \ref{fig3}(b). In other words, MBR consolidates the non-zero elements of a sparse matrix into a tighter central region around the diagonal and makes them easier to manage.

Based on the basic idea of MBR, this study implements a data schedule strategy according to the following steps:

\textbf{(1) Initialization}. A symmetrical adjacent matrix $M_{sym}$ is constructed from the original adjacent matrix $M_{ij}$, and an image id list ${list}_{id}$ is generated by using image indices that are deduced from rows or columns of $M_{ij}$.

\textbf{(2) Calculate permute order}. By using the MBR algorithm, a permute order ${order}_{perm}$ is generated, which permutes the row and column index of $M_{sym}$ and creates the MBR matrix $M_{mbr}$. This study adopts the Gibbs-Poole-Stockmeyer (GPS) algorithm \citep{gibbs1976algorithm} due to its low time costs and memory requirements. \added{For more details, refering to Appendex A.} In addition, the image id list ${list}_{id}$ is updated according to ${order}_{perm}$.

\textbf{(3) Create new schedule blocks}. By using MBR matrix $M_{mbr}$, new schedule blocks are created under the limitation of the GPU memory. As presented in Figure \mbox{\ref{fig3}(b)}, the GPU memory size ${Size}_{gpu}$ is restricted by the yellow grids, which indicates the maximum number of descriptor data that can be loaded into GPU memory. By defining the block size ${Size}_{blk}$, new schedule blocks ${{block}_{ij}}$ are created as rendered by red rectangles in Figure \mbox{\ref{fig3}(b)}, in which $i$ and $j$ represents the row and column index of each block, respectively. Block $\{{block}_{ij}\}$ stores its corresponding match pairs. For each block ${block}_{ij}$, images with their indices within the corresponding red rectangle are included, and their descirptor data would be loaded into GPU memory for feature matching.

\textbf{(4) Update adjacent matrix and image id list}. After feature matching guided by schedule blocks ${block}_{ij}$, the MBR matrix $M_{mbr}$ is updated by setting the cell value of the corresponding match pairs as zero and removing the rows and columns that have no match pairs. Besides, image id list ${list}_{id}$ is updated accordingly.

\textbf{(5) Termination}. Iteratively executing steps (2)-(4) until the MBR matrix $M_{mbr}$ is empty.

Compared with existing strategies, the proposed data schedule solution has two advantages. First, it minimizes data load and free in GPU as much as possible due to the usage of MBR for adjacent matric compression and block generation. Second, a high usage ratio of GPU can be achieved since match pairs as many as possible are packed into one schedule block.

\subsection{Cascade hashing-based feature matching}

\subsubsection{The workflow of cascade hashing feature matching}

Hashing-based feature matching converts real-value feature descriptors to binary hashing codes, and the similarity scores of feature descriptors can be calculated through a bit-wise XOR operation at a fast speed. In contrast to a single-layer structure, cascade hashing adopts a three-layer structure to implement coarse-to-fine feature matching \citep{cheng2014fast}, as shown in Figure \ref{fig2}. Because of the data-independent characteristic, LSH (Locality Sensitive Hashing) has been used as the hashing function in cascade hashing feature matching. Cascade hashing feature matching includes four steps:

\begin{figure}[!htp]
    \centering
    \includegraphics[width=1.0\linewidth]{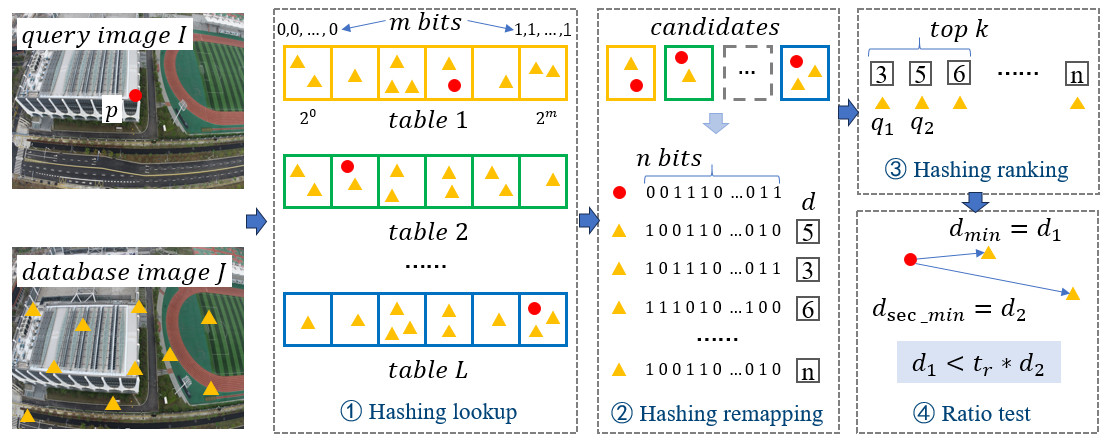}
    \caption{The workflow of cascade hashing-based feature matching.}
    \label{fig2}
 \end{figure}

\textbf{(1) Hashing lookup with multiple tables}. All feature descriptors in each image are mapped into $m$ bits short binary codes using LSH with $L$ sets of random hashing functions. For each set of hashing functions in $L$, a lookup table is created by using $2^m$ buckets. Thus, for query point $p$ in image $I$, all points that fall into the same bucket of $p$ in image $J$ are returned as its candidate matches.

\textbf{(2) Hashing remapping for fine encoding}. $m$ bits coarse hashing increases the recall of feature matching while the precision is low. Thus, candidate matches are remapped into $n\ (n>m)$ bits long binary codes with LSH. The Hamming distance that ranges from $0$ to $n$ is computed between the query point $p$ and their candidate matches, and it is used to sort candidate matches ascendly.

\textbf{(3) Hashing ranking for top-K nearest neighbors}. To achieve efficient NNS, hashing buckets with the Hamming distance as keys are established for all candidate matches. Nearest neighbors are returned by sequentially checking buckets until the number of returned matches reaches $K$.

\textbf{(4) Euclidean distance calculation and ratio test}. Hamming distances of hashing codes have lower precision than Euclidean distances of feature descriptors. For the top-K candidate matches, their Euclidean distances to the query point are calculated. The nearest neighbor $q_1$ and second nearest neighbor $q_2$ are determined, and $q_1$ that passes the ratio test $dis\left(q_1\right)<dis\left(q_2\right)\ast t_r$ is labeled as a final match. $t_r$ is the ratio test threshold, and $dis\left(\ast\right)$ indicates the distance to the query point.

According to cascade hashing feature matching, feature matches between images $I$ and $J$ can be established. As in \cite{cheng2014fast}, the default parameters are used in this study. That is, the number of tables $L=6$; the number of bits for coarse and fine encoding is set as $m=8$ and $n=128$, respectively; and the number of nearest neighbors is set as $K=8$. Noticeably, in ANN-based feature matching, the ratio test threshold $t_r$ balances the precision and recall of final matches, which is set as 0.8 in SIFTGPU. However, for cascade hashing matching, candidate matches have been selected in the coarse matching step (3). Thus, a proper $t_r$ would be evaluated.

\subsubsection{Schedule block guided cascade hashing feature matching}

Based on the principle of cascade hashing feature matching, the proposed solution is then implemented using the generated data schedule blocks $\{{block}_{ij}\}$. As the sum of data volume in each block row $i$ is restricted by GPU memory, feature descriptors in the block row can be loaded into GPU at one time. Supposing that generated blocks $\{{block}_{ij}\}$ consists of $k$ rows and $l$ columns, i.e., $i=1,2,\ldots,k$ and $j=1,2,\ldots,l$, and the current block row $i$ is represented as $\{{block}_{i1},{block}_{i2},\ldots,{block}_{il}\}$. The pipeline of cascade feature matching is implemented as follows:

\textbf{(1) Load feature descriptors from block}. Feature descriptors corresponding to match pairs in each block are sequentially loaded into GPU, and their hashing codes and bucket ids are simultaneously computed to build hashing lookup tables.

\textbf{(2) Execute cascade hashing matching}. Based on the established hashing lookup tables, hashing feature matching is executed and guided by match pairs, and initial matches that pass through the ratio test are retained.

\textbf{(3) Free unnecessary data block}. Iteratively execute steps (1) and (2) until all blocks in the current block row are processed. The data in the first block ${block}_{i1}$ is free as it would not be used in the next block row.

For each block row $\{{block}_{i1},{block}_{i2},\ldots,{block}_{il}\}$, the above pipeline is executed to obtain initial candidate matches for all involved match pairs. Noticeably, in step (1), redundant data loading operations can be avoided. As illustrated in Figure \ref{fig3}(b) for processing the second block row, only descriptors in the last block ${block}_{il}$ rendered in orange color should be loaded into GPU since all the other required data has been loaded when processing the first block row. Besides, after processing each block row, data in the first block ${block}_{i1}$ rendered in green color is free as in step (3). The interleave free-load operation goes through the whole pipeline.

\subsubsection{Outlier removal combining local and global geometric constraints}

\begin{figure}[!htp]
    \centering
    \includegraphics[width=1.0\linewidth]{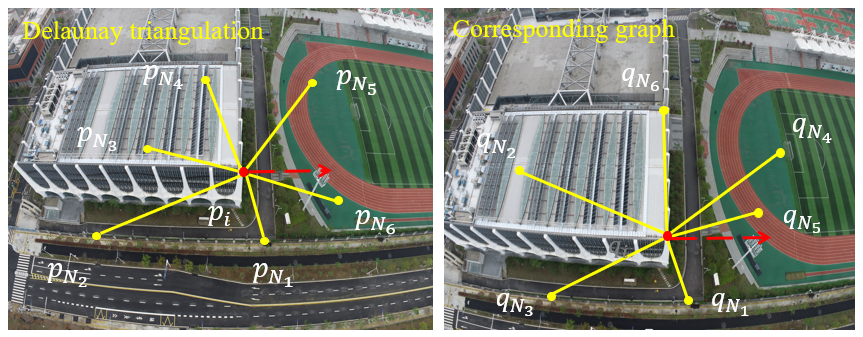}
    \caption{The principle of the SAO-based local geometric constraint.}
    \label{fig4}
 \end{figure}

False matches inevitably existed in initial matches due to limited discriminability of local feature descriptors and large perspective deformations. In this study, outlier removal is finally conducted by combining the spatial angular order (SAO) based local geometric constraint and RANSAC-based global geometric constraint \citep{jiang2020reliable}. The local geometric constraint is designed by using the k-nearest neighbors deduced from the Delaunay triangulation and its corresponding graph, as shown in Figure \mbox{\ref{fig4}}. For one candidate match ${p_i, q_i}$, their K nearest neighbors $p_{N_i}$ and $q_{N_i}$ are first found and ordered based on their angular order with respect to the horizonal axis. For example, the K nearest neighbors of $p_i$ are $O_{p_i}={{p_{N_5}, p_{N_4}, p_{N_3}, p_{N_2}, p_{N_1}, p_{N_6}}}$. Similarly, the K nearest neighbors $O_{q_i}$ of $q_i$ can also be found. The dissimilarity score $Score$ of the candidate match ${p_i, q_i}$ is then calcualted by using the cyclic edit distance (CED) algorithm between these two sets of nearest neighbors, i.e., $Score=\frac{1}{N}d_{ced}(O_{p_i}, O_{q_i})$. $N$ is the number of neighbors; $d_{ced}(.)$ calculates the CED. To achieve further efficiency improvement, outlier removal is simultaneously executed in CPU after obtaining initial matches in step (2), in which initial matches with larger dissimilarity scores are prone to false matches. Thus, the results retained in outlier removal are the final matches.

\subsection{Algorithm implementation}

According to the workflow presented in Figure \ref{fig1}, this study implements the proposed feature matching algorithm by using the C++ programming language. For match pair retrieval, SIFTGPU \citep{wu2011siftgpu} with default parameters is utilized to extract SFIT features, and Lloyd’s K-means cluster algorithm \citep{lloyd1982least} is adopted to cluster selected training descriptors for the codebook construction. With the aid of VLFeat \citep{vedaldi2010vlfeat} and FAISS packages \citep{johnson2019billion}, we have implemented the VLAD-HNSW based image retrieval solution. To facilitate performance evaluation in SfM reconstruction, the cascade hashing feature matching solution has been integrated into the SfM engine in \cite{jiang2022parallel}. The pseudo-code is presented in Algorithm \ref{alg1}. \added{Noticeable, in the procedure CascadeHashingMatching, lines 4 to 8 for MBR-based schedule block generation are executed in CPU, and lines 10 to 14 uses both GPU and CPU for feature matching and outlier removal. These two steps operate between CPU and GPU. The symbols $\uparrow$ and $\downarrow$ indicate uploading data from host memory to GPU memory and releasing data in GPU, respectively. The symbols $\equiv$ and $-$ indicate feature matching in GPU and outlier removal in CPU, respectively. These fives steps are executed asynchronously.}

\begin{algorithm}[!t]
    \caption{cascade hashing feature matching}
    \textbf{Input:} UAV images $I=\{i_i\}$ of size $n$ \\
    \textbf{Output:} Feature matching results $match=\{m_{ij}\}$
    \begin{algorithmic}[1]
    \Procedure{ViewGraphConstruction}{}
      \State Extract SIFT features for each image $i_i \in I$
      \State Select a subset of $p$ percent of images from $I$ as training images $I_p$
      \State Add top-scale $h$ features of image $i_i \in I_p$ to training descriptors $D_t$
      \State Generate a codebook $C$ with $k$ words using training descriptors $D_t$
      \State Retrieve match pairs $P=\left\{p_{ij}\right\}$ via VLAD-HNSW
      \State Create View graph $M_{ij}$ using selected match pairs $P$
    \EndProcedure
    \end{algorithmic}
    \begin{algorithmic}[1]
    \Procedure{CascadeHashingMatching}{}
      \State Initialize symmetrical matrix $M_{sym}$ and image id list ${list}_{id}$ from $M_{ij}$
      \While{$M_{sym}$ is not empty}
        \State Assign schedule blocks $block=\{\}$
        \State Calculate permute order ${order}_{perm}$ from symmetrical matrix $M_{sym}$
        \State Create MBR matrix $M_{mbr}$ and update list ${list}_{id}$ using ${order}_{perm}$
        \State Create schedule blocks from $M_{mbr}$ and add to ${block}\gets\{{block}_{ij}\}$
        \State Update symmetrical matrix $M_{sym}=M_{mbr}$
        
        \ForEach{$\{{block}_{i1},{block}_{i2},\ldots,{block}_{il}\} \in block$}
          \State \textcolor{blue}{$\uparrow$} Load descriptors in current block ${block}_{ij}$
          \State \textcolor{blue}{$\equiv$} Execute cascade hashing matching for ${block}_{ij}$
          \State \textcolor{blue}{$-$} Remove outliers and store results $match\gets\left\{m_{ij}\right\}$
          \State \textcolor{blue}{$\downarrow$} Free unnecessary data in block ${block}_{i1}$
          \State \textcolor{blue}{$-$} Set matrix $M_{sym}$ cells as zero for processed pairs
        \EndFor
      \EndWhile
    \EndProcedure
    \end{algorithmic}
    \label{alg1}
\end{algorithm}

\section{Experimental results}
\label{sec4}

Three large-scale UAV datasets are utilized for performance evaluation. First, we analyze the influence of the ratio test threshold and schedule block size on feature matching precision and efficiency. Second, the key steps of the proposed solution are evaluated using ablation studies of feaure matching. Finally, we compare the proposed algorithm with open-source and commercial software packages in terms of feature matching and SfM-based 3D reconstruction. In this study, all algorithms have been implemented by using the C++ programming language, and the tests are conducted on a Windows platform equipped with 64 GB memory, a 3.0 GHz Intel Core i9-13900K CPU and a 24 GB NVIDIA GeForce RTX 4090 graphic card.

\subsection{Datasets and software packages}
\label{sec4.1}

Three UAV datasets have been collected in study for performance evaluation. The details of data acquisitions are presented in Table \ref{tab1}.

\begin{table}[!t]
	\centering
	\caption{Detailed information about the three UAV datasets.}
	\label{tab1}
	\makebox[1\linewidth]{
		\begin{tabular}{l l l l}
			\toprule
			\textbf{Item Name} & \textbf{Dataset 1} & \textbf{Dataset 2} & \textbf{Dataset 3} \\
			\midrule
			UAV type & multi-rotor &  multi-rotor & multi-rotor \\
			Flight height (m) & 80.0 & - & 87.0 \\
			Camera mode & DJI FC6310R & DJI ZenmuseP1 & SONY ILCE 7R \\
			Number of cameras & 1 & 1 & 5 \\
			Focal length (mm) & 24 & 35 & 35 \\
			\multirow{2}{*}{Camera angle ($^\circ$)} & \multirow{2}{*}{0} & \multirow{2}{*}{-} & \multirow{2}{*}{\makecell[l]{Nadir: 0; \\
                oblique: 45/-45}} \\
                &&&\\
			Number of images & 3,743 & 4,030 & 21,654 \\
			Image size (pixel)  & 5472 $\times$ 3648 & 8192 $\times$ 5460 & 6000 $\times$ 4000 \\
			GSD (cm) & 2.6 & 1.2 & 1.2 \\
			\bottomrule
		\end{tabular}
	}
\end{table}

\begin{figure}[!htp]
    \centering
    \begin{minipage}[t]{0.8\linewidth}
    \centering
        \begin{tabular}{@{\extracolsep{\fill}}c@{}c@{}c@{}@{\extracolsep{\fill}}}
            \includegraphics[width=0.7\linewidth]{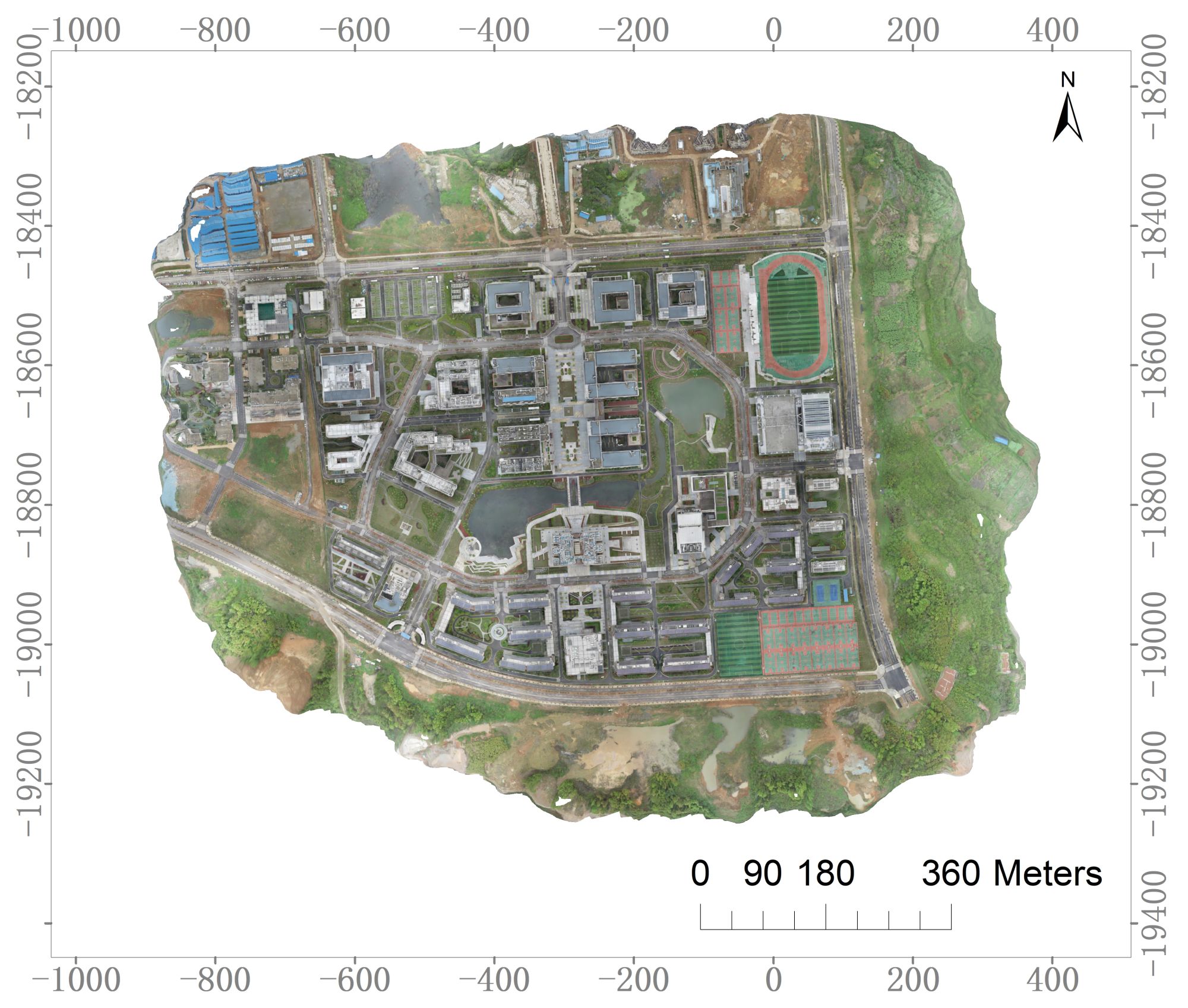} \\
            (a) Dataset 1  \\
        \end{tabular}
    \end{minipage}
    \begin{minipage}[t]{0.8\linewidth}
    \centering
        \begin{tabular}{@{\extracolsep{\fill}}c@{}c@{}@{\extracolsep{\fill}}}
            \includegraphics[width=0.7\linewidth]{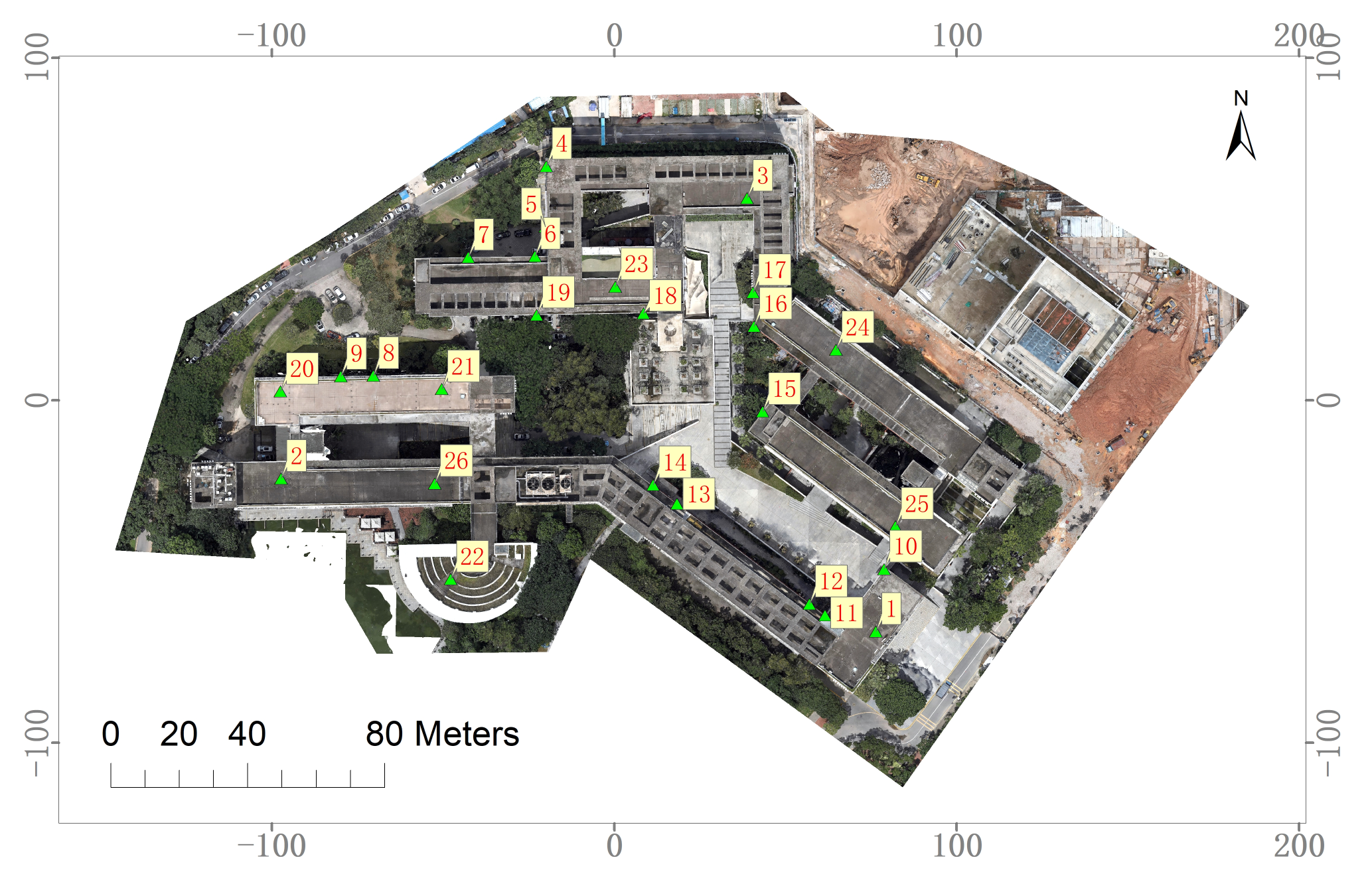} \\
            (b) Dataset 2 \\
        \end{tabular}
    \end{minipage}
    \begin{minipage}[t]{0.8\linewidth}
    \centering
        \begin{tabular}{@{\extracolsep{\fill}}c@{}c@{}@{\extracolsep{\fill}}}
            \includegraphics[width=0.7\linewidth]{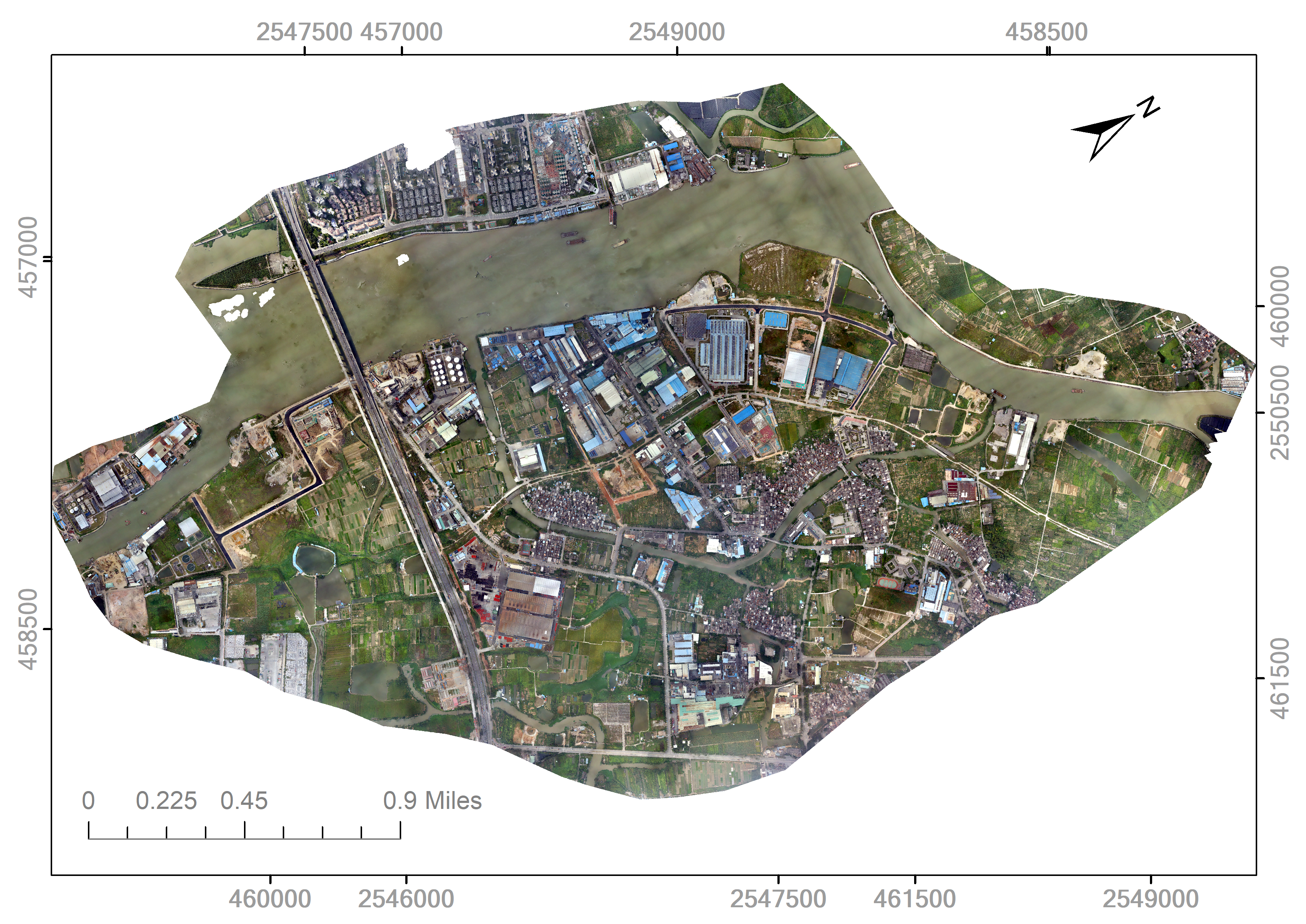} \\
            (c) Datasets 3 \\
        \end{tabular}
    \end{minipage}
    \caption{The orthomosaics of the three test sites.}
    \label{fig5}
 \end{figure}

\begin{itemize}
    \item The first dataset comprises 3,743 images over a university campus, as presented in Figure \ref{fig5}(a). These images are obtained by using a DJI Phantom 4 RTK UAV, which is equipped with a DJI FC6310R camera. Each image has 5,472 by 3,648 pixels and is taken at a flight altitude of 80 m, yielding a Ground Sample Distance (GSD) of about 2.6 cm.
    \item The second dataset encompasses 4,030 images from a complex university building as shown in Figure \ref{fig5}(b). These images are captured by a DJI M300 RTK UAV, equipped with a DJI Zenmuse P1 camera that produces images with dimensions of 8,192 by 5,460 pixels. The dataset has been collected using optimized views photogrammetry \citep{li2023optimized}, which adjusts camera orientation according to geometric attributes of terrestrial objects. The GSD is about 1.2 cm. For absolute orientation, 26 Ground Control Points (GCPs) have been gathered using a total station, which offers a nominal precision of 0.8 cm and 1.5 cm in the horizontal and vertical direction, respectively. 
    \item The third dataset is captured from a test site that is characterized by low-rise structures and an abundance of dense vegetation, with a river flowing through the area as shown in Figure \ref{fig5}(c). For data acquisition, a typical penta-view oblique photogrammetry with five SONY ILCE 7R cameras has been adopted, which record images with dimensions of 6,000 by 4,000 pixels. At an altitude of 87.1 m, a collection of 21,654 images is gathered, achieving a GSD of 1.2 cm.
\end{itemize}

\begin{table*}[!t]
	\centering
	\caption{The configuration of the evaluated methods. The term VOC-Tree indicates vocabulary tree-based image retrieval.}
	\label{tab2}
	\makebox[1\linewidth]{
    \begin{tabular}{l l l l l l}
                \toprule
                \multirow{2}{*}{\textbf{Method}} & \multirow{2}{*}{\makecell[l]{\textbf{Match pair} \\ \textbf{selection}}} & \multirow{2}{*}{\makecell[l]{\textbf{Feature matching} \\ \textbf{algorithm}}} & \multirow{2}{*}{\makecell[l]{\textbf{Data} \\ \textbf{schedule}}} & \multirow{2}{*}{\makecell[l]{\textbf{Hardware} \\ \textbf{acceleration}}} & \multirow{2}{*}{\makecell[l]{\textbf{Software} \\ \textbf{version}}} \\
                &  &  &  &  &  \\
                \midrule
                ColMap-CPU     & Input pairs     & KD-Tree          & No  & Multi-core CPU  & 3.10 \\
                ColMap-GPU     & Input pairs     & NNS              & Yes & GPU \& CPU      & 3.10 \\
                AliceVision    & Input pairs     & Cascade hashing  & No  & Multi-core CPU  & v2023.3.0 \\
                TLBDS          & Input pairs     & Cascade hashing  & Yes & GPU \& CPU      & / \\
                GraphPart          & Input pairs     & Cascade hashing  & Yes & GPU \& CPU      & / \\
                Metashape      & POS             & /                & Yes & GPU \& CPU      & 1.8.21 \\
                Pix4Dmapper    & VOC-Tree        & /                & Yes & GPU \& CPU      & 4.4.12 \\
                Ours           & Input pairs     & Cascade hashing  & Yes & GPU \& CPU      & / \\
                \bottomrule
		\end{tabular}
	}
\end{table*}

For performance evaluation, this study compares the proposed algorithm with six well-known solutions, as presented in Table \mbox{\ref{tab2}}, including two open-source software packages ColMap \mbox{\citep{schonberger2016structure}} and AliceVision \mbox{\citep{griwodz2021alicevision}}, two commercial software packages Metashape \mbox{\citep{Metashape}} and Pix4Dmapper \mbox{\citep{Pix4dMapper}}, one classical graph partition based matching algorithm GraphPart, and one recently cascade hashing matching algorithm TLBDS \mbox{\citep{zhang2023efficient}}. GraphPart is implmented by splitting the whole view graph into sub-graphs with pre-defined vertex number.

\subsection{Analysis of schedule block size and ratio test threshold}
\label{sec4.2}

In the proposed feature matching algorithm, the schedule block size ${Size}_{blk}$ and ratio test threshold $t_r$ are two critical parameters. The schedule block size ${Size}_{blk}$ determines the data quantity that is loaded while processing each block row and the count of match pairs that are fed to GPU. For a small block size, the load and free of data are consistently less demanding, which in turn results in reduced GPU usage. Conversely, a large block size can process more matching pairs simultaneously, thereby increasing GPU usage and data scheduling demands. The ratio test threshold $t_r$ determines the number of initial matches that are retained in the cascade hashing feature matching. In contrast to global NNS, candidate matches have been filtered in the coarse matching step, which leads to the practical threshold used in the other algorithm being improper. Thus, we will analyze their influence on feature matching and select their optimal values.

\begin{figure}[!tp]
    \centering
    \includegraphics[width=0.8\linewidth]{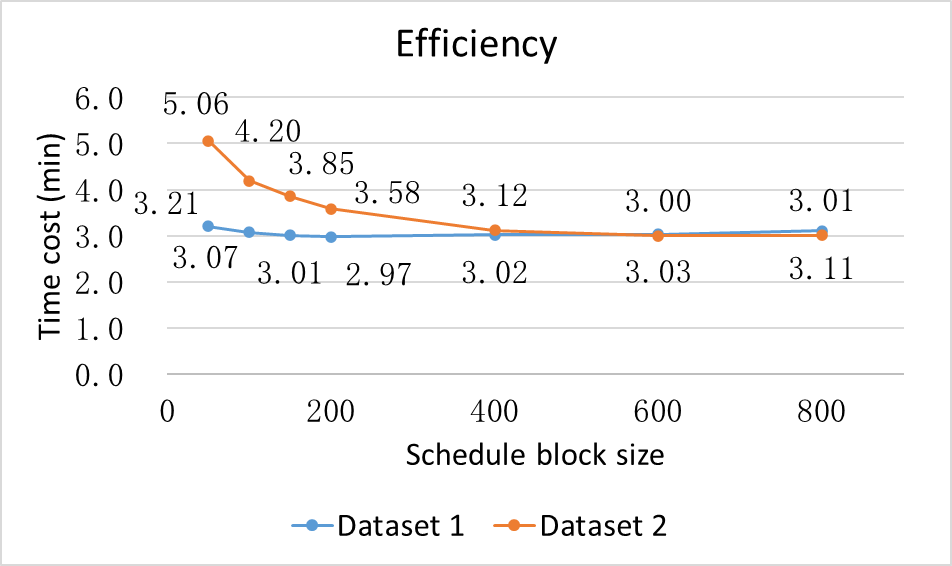}
    \caption{The influence of the schedule block size on matching efficiency.}
    \label{fig6}
 \end{figure}

For the analysis of schedule block size ${Size}_{blk}$, datasets 1 and 2 are selected as they are captured by using two different photogrammetric configurations, which results in the collected images with varying viewing directions and image overlap degrees. In this test, the schedule block size is sampled with the values of 50, 100, 150, 200, 400, 600, and 800 under the limitation of the maximum GPU memory, and time cost is used as the metric for performance evaluation. The results are shown in Figure \ref{fig6}. It is shown that the time cost decreases consistently with the increase of schedule block size ranging from 50 to 400. The main reason is that for the large block size, more match pairs are fed to GPU at one time, which would increase the GPU usage and in turn increase matching efficiency. Noticeably, higher efficiency improvement is observed from dataset 2 when compared with that in dataset 1. It can be explained by the larger image overlap degree in dataset 2. When the schedule block size further increases from 400 to 800, it is shown that the time cost almost keeps constant for these two datasets due to the full usage of the GPU computing power. Thus, ${Size}_{blk}$ is set as 400 in following tests.

 \begin{figure}[!t]
    \centering
    \begin{minipage}[t]{0.8\linewidth}
    \centering
        \includegraphics[width=1\linewidth]{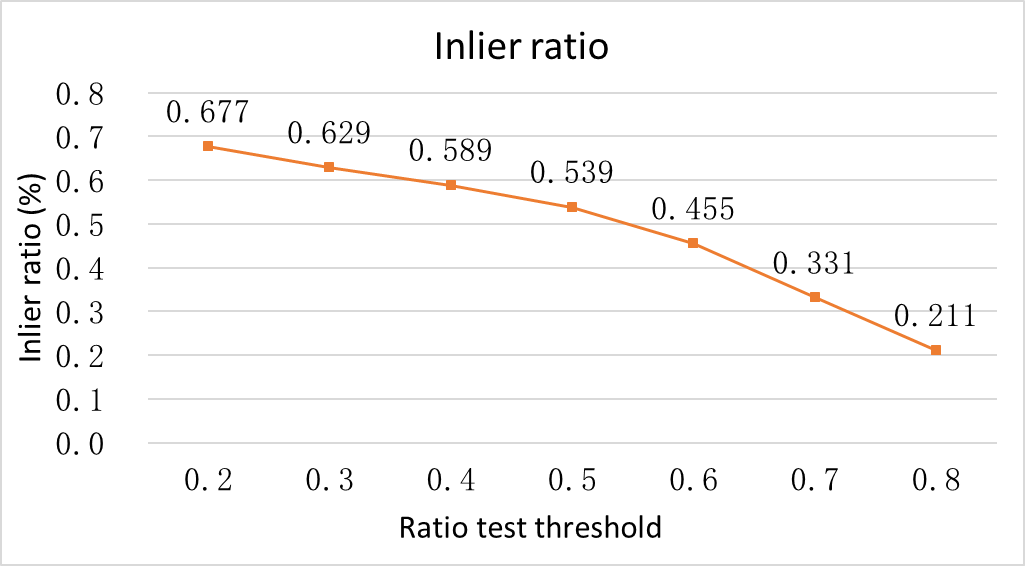} \\
        (a)  \\
    \end{minipage}
    \begin{minipage}[t]{0.8\linewidth}
    \centering
        \includegraphics[width=1\linewidth]{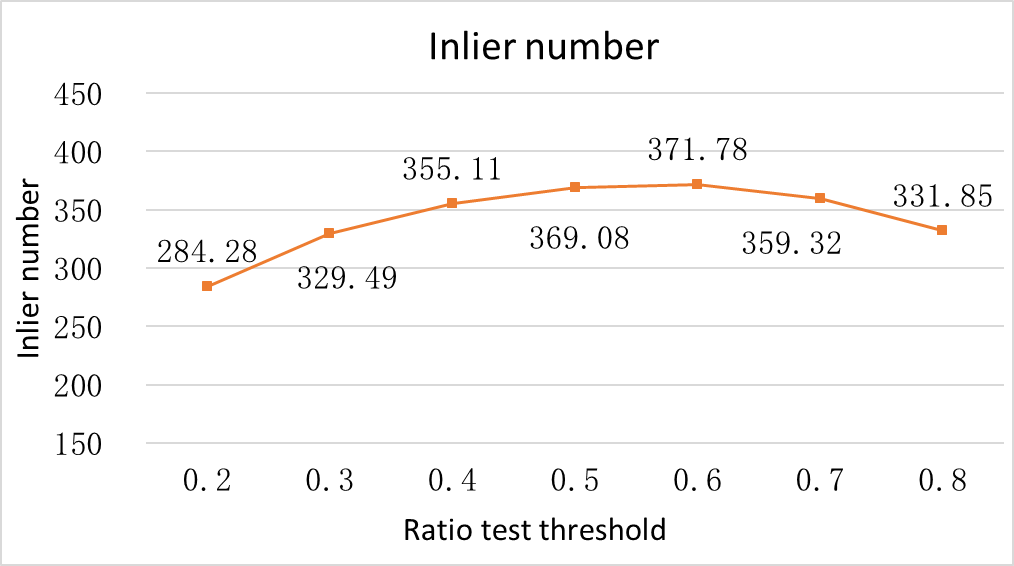} \\
        (b) \\
    \end{minipage}
    \caption{The influence of the ratio test threshold on inlier ratio and inlier number using dataset 2: (a) average inlier ratio; and (b) average inlier number.}
    \label{fig7}
\end{figure}

\begin{figure}[!tp]
    \centering
        \begin{minipage}[t]{1.0\linewidth}
    \centering
        \begin{tabular}{@{\extracolsep{\fill}}c@{}c@{}c@{}@{\extracolsep{\fill}}}
            \includegraphics[width=1\linewidth]{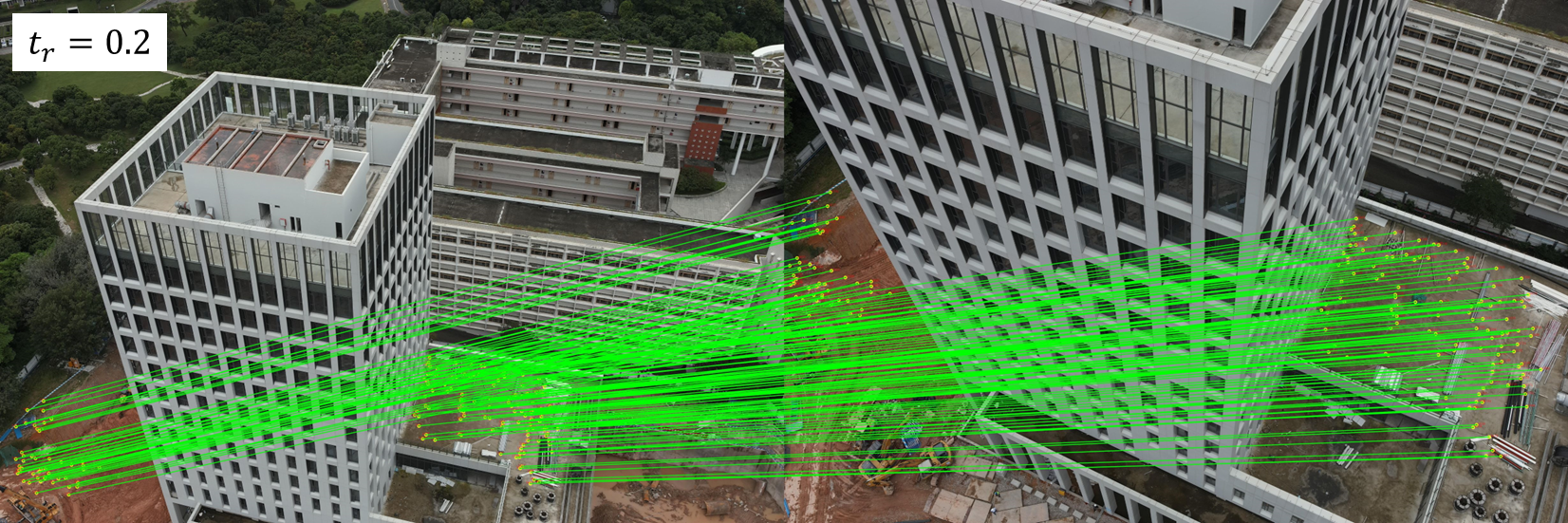}\\
            (a) 183/381/0.48 \\
        \end{tabular}
    \end{minipage}
    \begin{minipage}[t]{1.0\linewidth}
    \centering
        \begin{tabular}{@{\extracolsep{\fill}}c@{}c@{}c@{}@{\extracolsep{\fill}}}
            \includegraphics[width=0.5\linewidth]{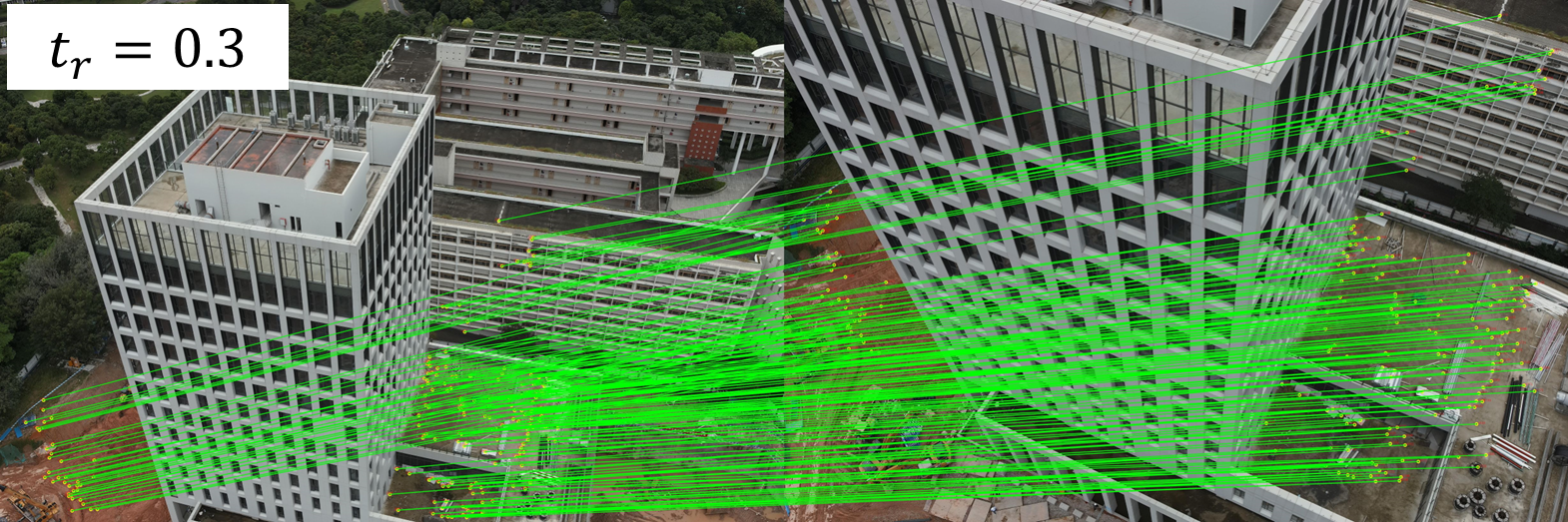} &
            \includegraphics[width=0.5\linewidth]{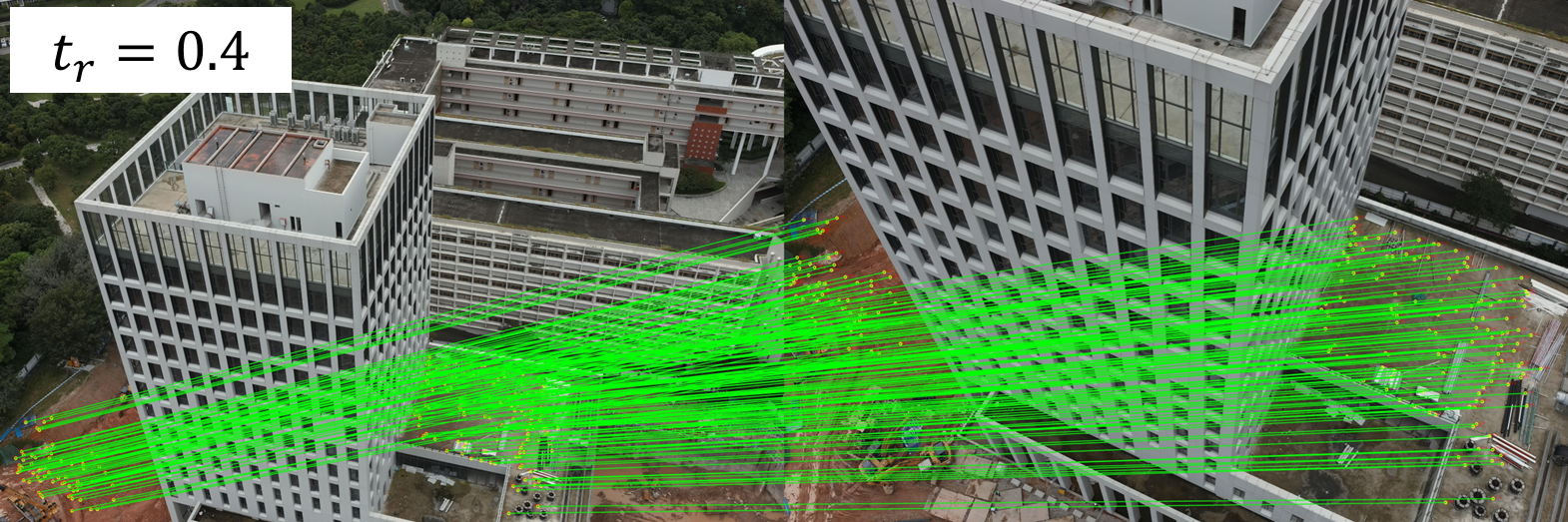}\\
            (b) 211/447/0.47 & (c) 218/510/0.43 \\
        \end{tabular}
    \end{minipage}
    \begin{minipage}[t]{1.0\linewidth}
    \centering
        \begin{tabular}{@{\extracolsep{\fill}}c@{}c@{}@{\extracolsep{\fill}}}
            \includegraphics[width=0.5\linewidth]{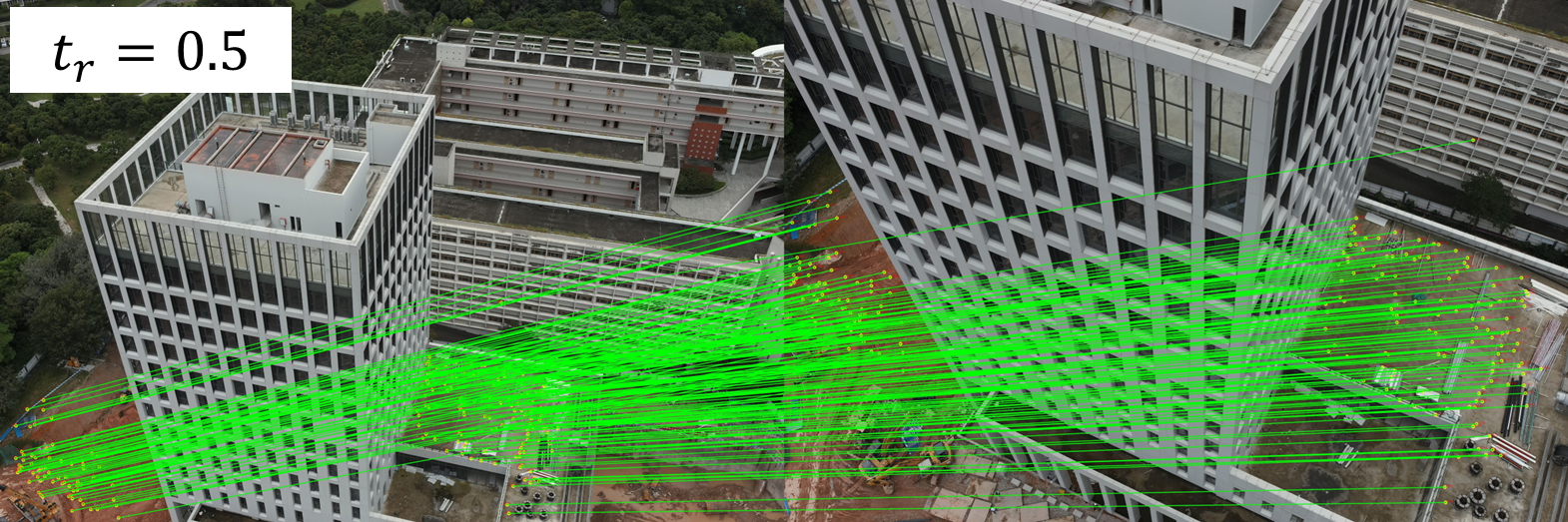} &
            \includegraphics[width=0.5\linewidth]{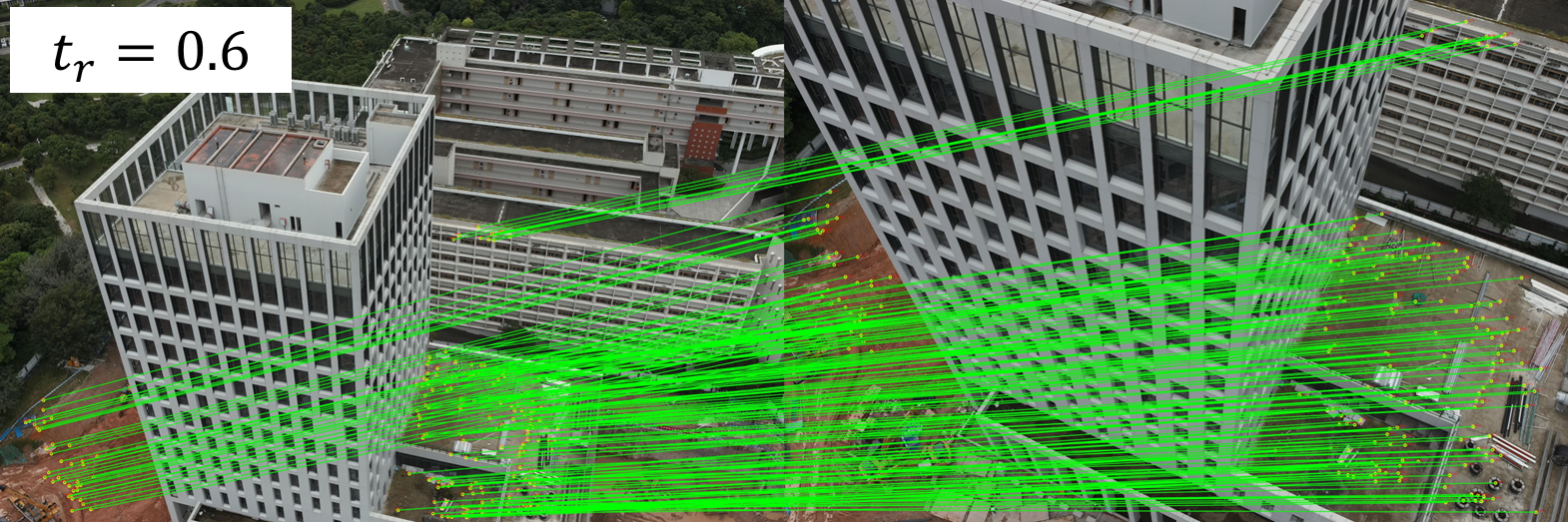}\\
            (d) 213/583/0.37 & (e) 226/677/0.33 \\
        \end{tabular}
    \end{minipage}
    \begin{minipage}[t]{1.0\linewidth}
    \centering
        \begin{tabular}{@{\extracolsep{\fill}}c@{}c@{}@{\extracolsep{\fill}}}
            \includegraphics[width=0.5\linewidth]{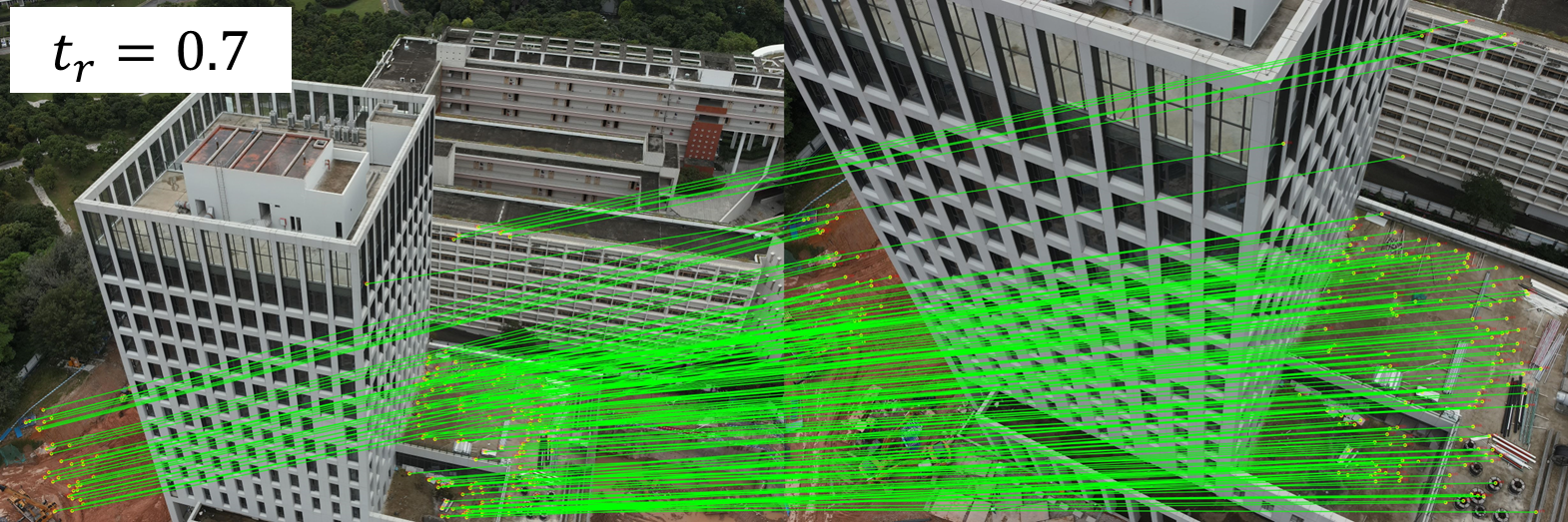} &
            \includegraphics[width=0.5\linewidth]{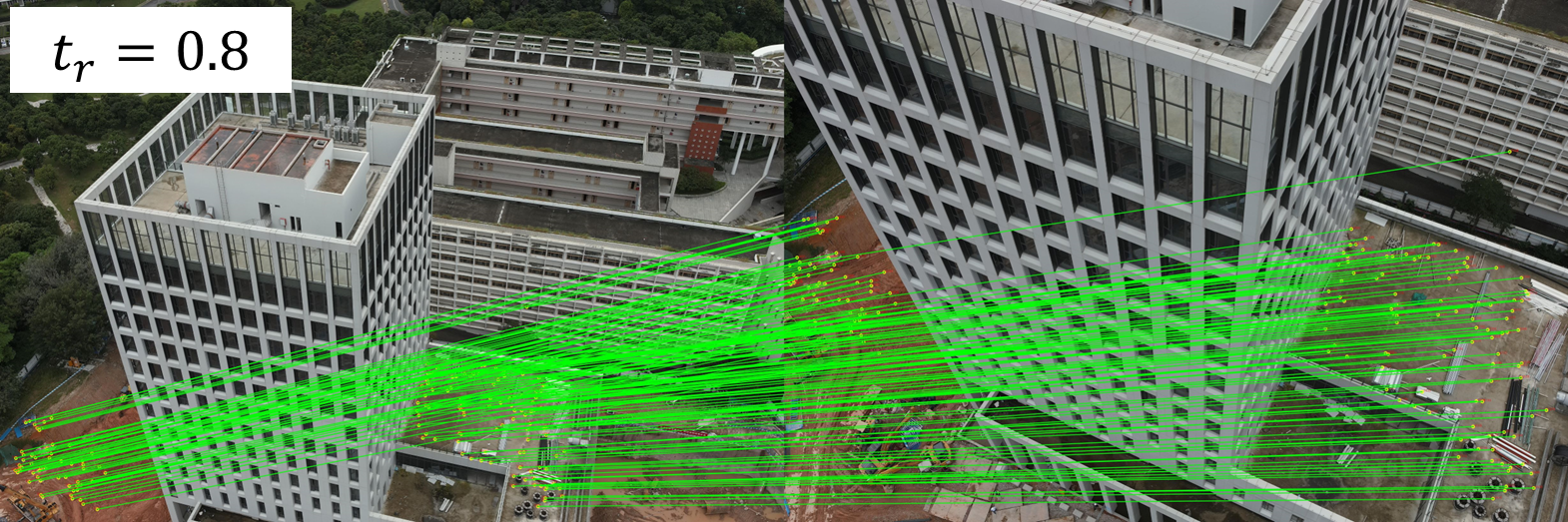}\\
            (f) 220/868/0.25 & (g) 182/1224/0.15 \\
        \end{tabular}
    \end{minipage}
    \caption{Feature matching results of one image pair in dataset 2 under different ratio test threshold $t_r$. The values along with the title of each sub-figure indicate the number of inliers and initial matches and its corresponding inlier ratio. Noticeably, presented results are final matches after outlier removal.}
    \label{fig8}
 \end{figure}

For the analysis of ratio test threshold $t_r$, dataset 2 is selected for tests due to the existence of different and classical structures. In this test, the ratio test threshold is sampled from 0.2 to 0.8 with an interval value of 0.1. The inlier ratio and inlier number are used as the metrics for performance evaluation. The former is the ratio between the number of inliers and all matches; the latter is the number of true matches after outlier removal. Figure \ref{fig7} presents the results. We can see that with the increase of ratio test threshold, the inlier ratio continuously decreases from 0.677 to 0.211, as shown in Figure \ref{fig7}(a). The main reason is that more outliers pass the ratio test with the increase of the ratio test threshold when compared with the number of inliers. For the results in Figure \ref{fig7}(b), the change of the inlier number can be divided into two stages. In the first stage, it increases from 284.28 to 371.78 for the threshold increasing from 0.2 to 0.6. The main reason is that more matches pass the ratio test, which in turn increases the inlier number. On the second stage ranging from 0.6 to 0.8, the inlier number continuously decreases as the lower inlier ratio degenerates the performance of subsequent outlier removal. For a visual analysis, Figure \ref{fig8} presents the feature matching result of one image pair under varying threshold. Thus, for a balance between inlier ratio and inlier number, $t_r$ is set as 0.5 in the following test.

\subsection{Analysis of the key steps of the proposed feature matching solution}
\label{sec4.3}

\subsubsection{Image retreival-based match pair selection}
\label{sec4.3.1}

\begin{figure}[!t]
    \centering
    \begin{minipage}[t]{1.0\linewidth}
    \centering
        \begin{tabular}{@{\extracolsep{\fill}}c@{}c@{}c@{}@{\extracolsep{\fill}}}
            \includegraphics[width=0.33\linewidth]{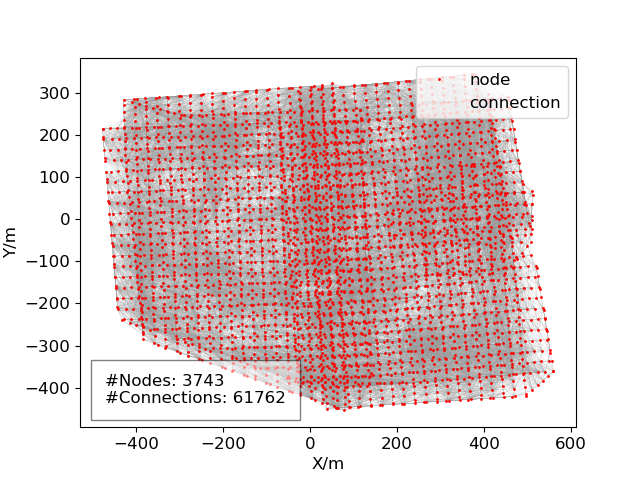} &
            \includegraphics[width=0.33\linewidth]{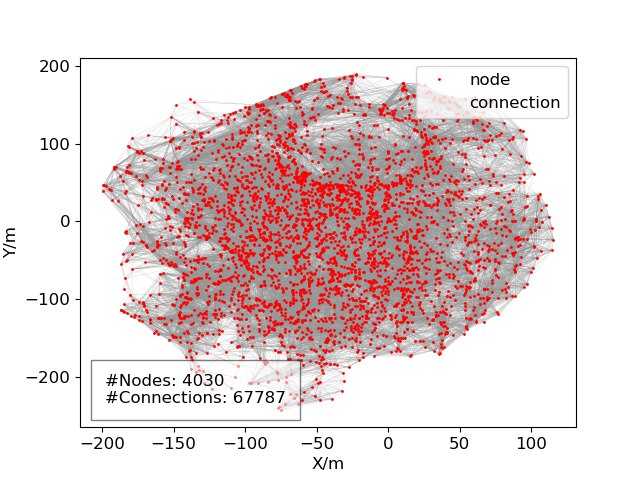} &
            \includegraphics[width=0.33\linewidth]{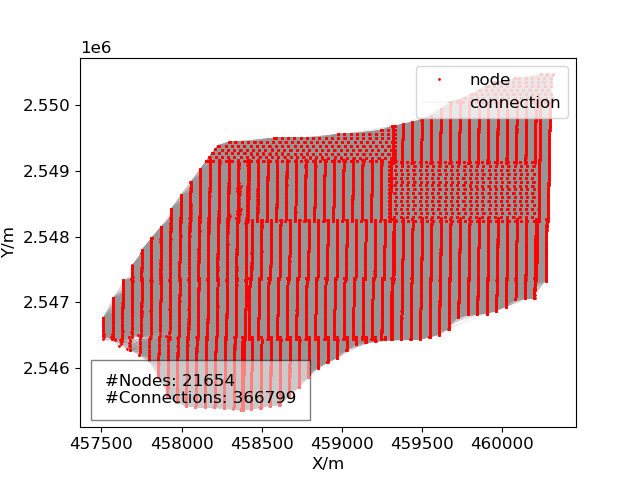}\\
            (a) & (b) & (c)\\
        \end{tabular}
    \end{minipage}
   
    \caption{Image connection network created from match pairs: (a) dataset 1; (b) dataset 2; (c) dataset 3. Red dots and gray lines respectively indicate images and connections.}
    \label{fig9}
 \end{figure}

Match pair selection is the first step of the proposed feature matching workflow. In this study, the HNSW-VLAD based image retrieval technique has been used to achieve efficient match pair selection for the three datasets. Three metrics, i.e., \textit{efficiency}, \textit{precision}, and the \textit{number of match pairs}, are used for performance evaluation. The first one indicates the time cost in image retrieval; the second one is the ratio between the number of true match pairs and all retrieved match pairs. The statistical results are presented in Table \ref{tab3}. We can conclude that for the three datasets there are a total number of 61,762, 67,787, and 366,799 match pairs selected under the time costs of 1.0 min, 1.2 min, and 8.8 min, respectively. In other words, the average time cost is almost independent of the data volume, which is approximately 0.97 ms, 1.06 ms, and 1.44 ms for the three datasets, respectively. In addition, the metric \textit{precision} for the three datasets is 89.8\%, 89.0\%, and 96.0\%, which ensures high precision for the retrieved match pairs to create the subsequent view graph, as shown in Figure \ref{fig9}. Thus, the proposed algorithm can easily scale to very large-scale UAV datasets.

\begin{table}[!t]
	\centering
	\caption{The statistical results of match pair selection for the three datasets in terms of efficiency, precision, and the number of match pairs.}
	\label{tab3}
	\makebox[0.5\linewidth]{
		\begin{tabular}{l l l l}
			\toprule
			\textbf{Metric} & \textbf{Dataset 1} & \textbf{Dataset 2} & \textbf{Dataset 3} \\
			\midrule
			Efficiency (min)  &	1.0 &	1.2 &	8.8  \\
			Precision (\%) &	89.8&	89.0&	96.0\\
                Number of match pairs&	61,762	&67,787	&366,799\\
			\bottomrule
		\end{tabular}
	}
\end{table}

\begin{figure}[!t]
    \centering
    \begin{minipage}[t]{1.0\linewidth}
    \centering
        \begin{tabular}{@{\extracolsep{\fill}}c@{}c@{}c@{}@{\extracolsep{\fill}}}
            \includegraphics[width=0.3\linewidth]{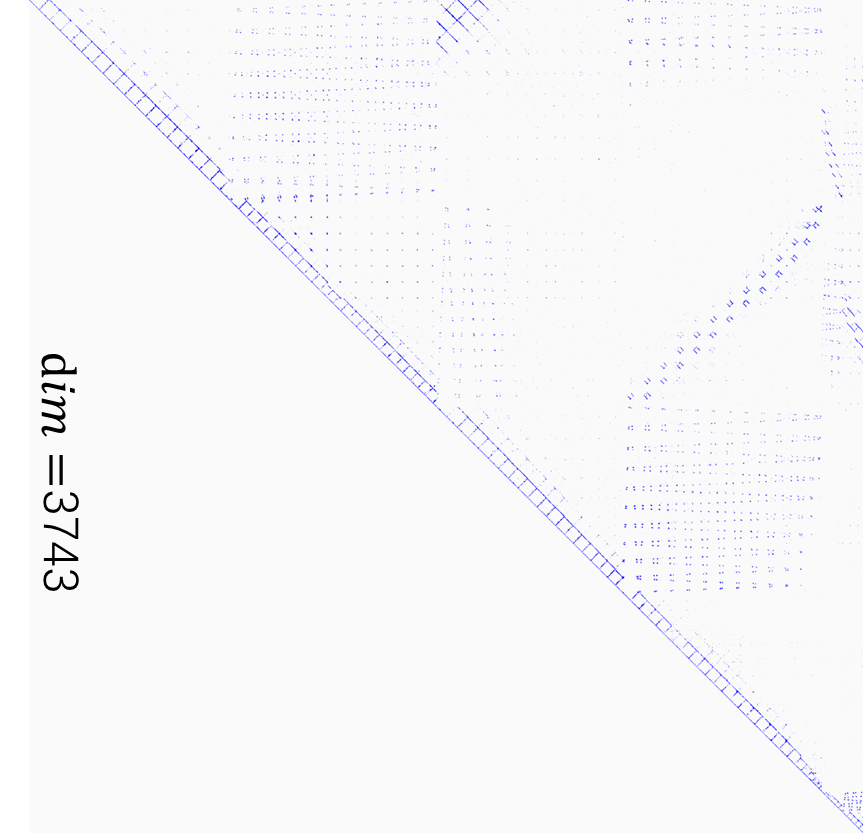} &
            \includegraphics[width=0.3\linewidth]{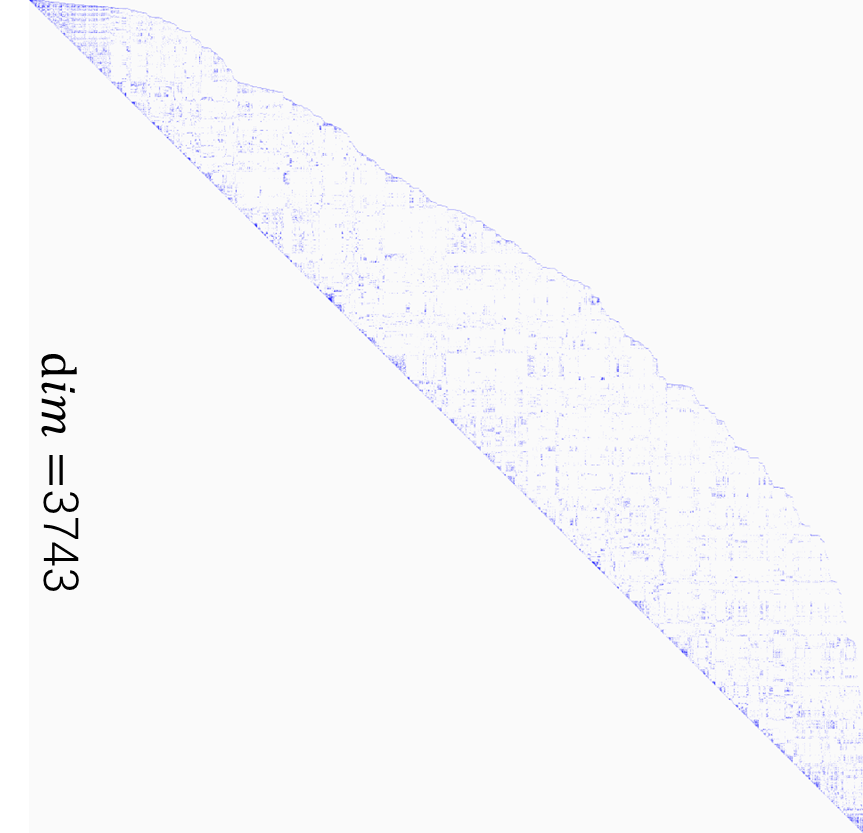} &
            \includegraphics[width=0.3\linewidth]{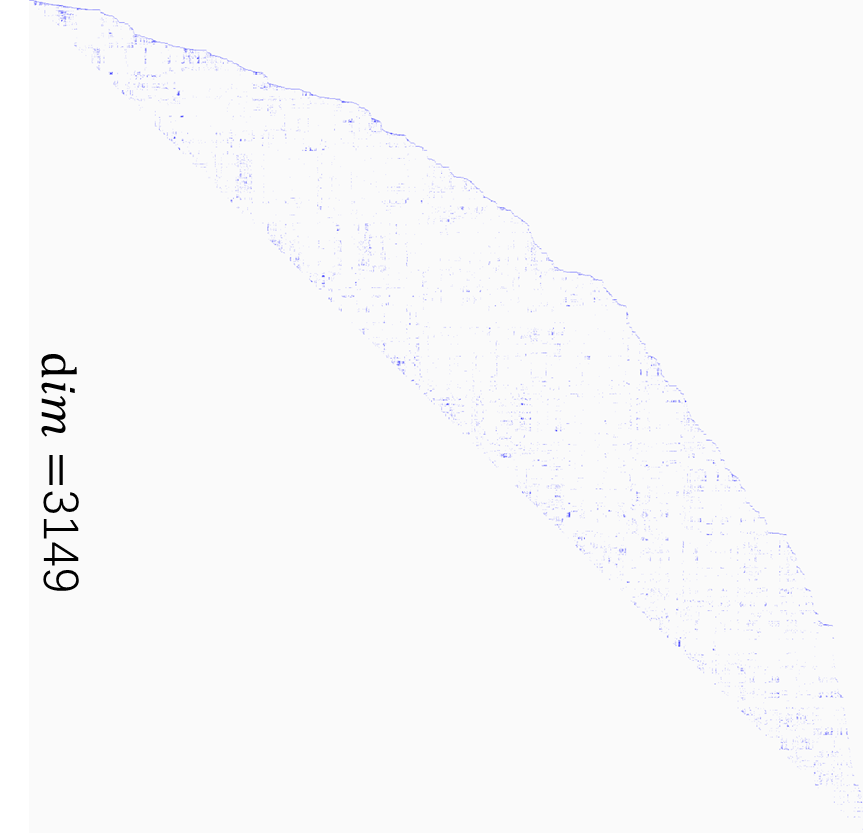}\\
            (a) & (b) & (c)\\
        \end{tabular}
    \end{minipage}

    \begin{minipage}[t]{1.0\linewidth}
    \centering
        \begin{tabular}{@{\extracolsep{\fill}}c@{}c@{}c@{}@{\extracolsep{\fill}}}
            \includegraphics[width=0.3\linewidth]{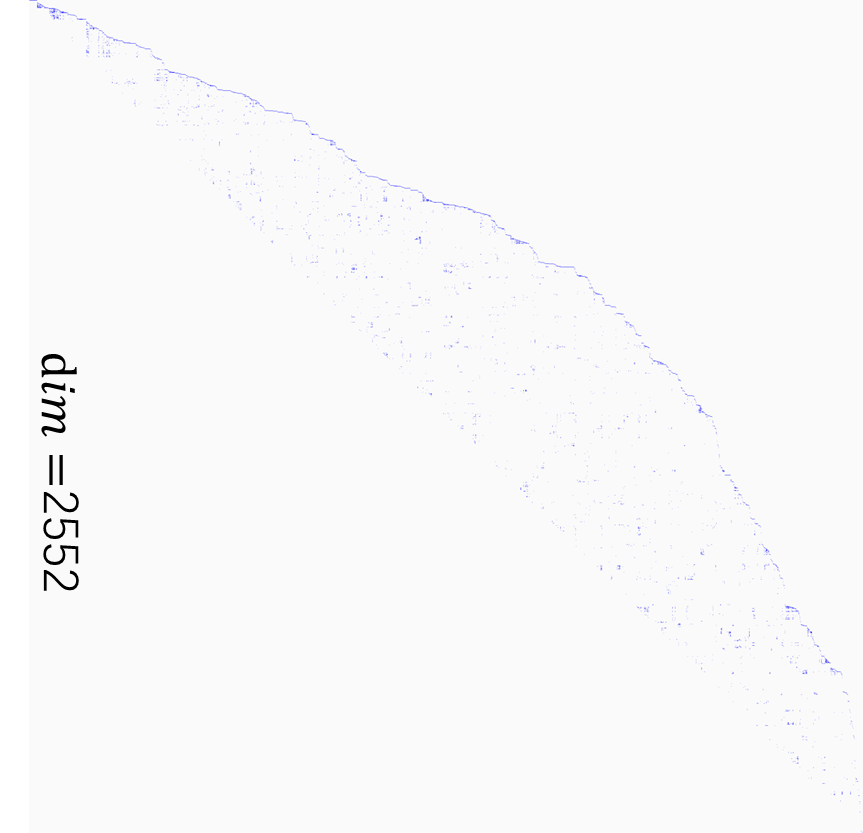} &
            \includegraphics[width=0.3\linewidth]{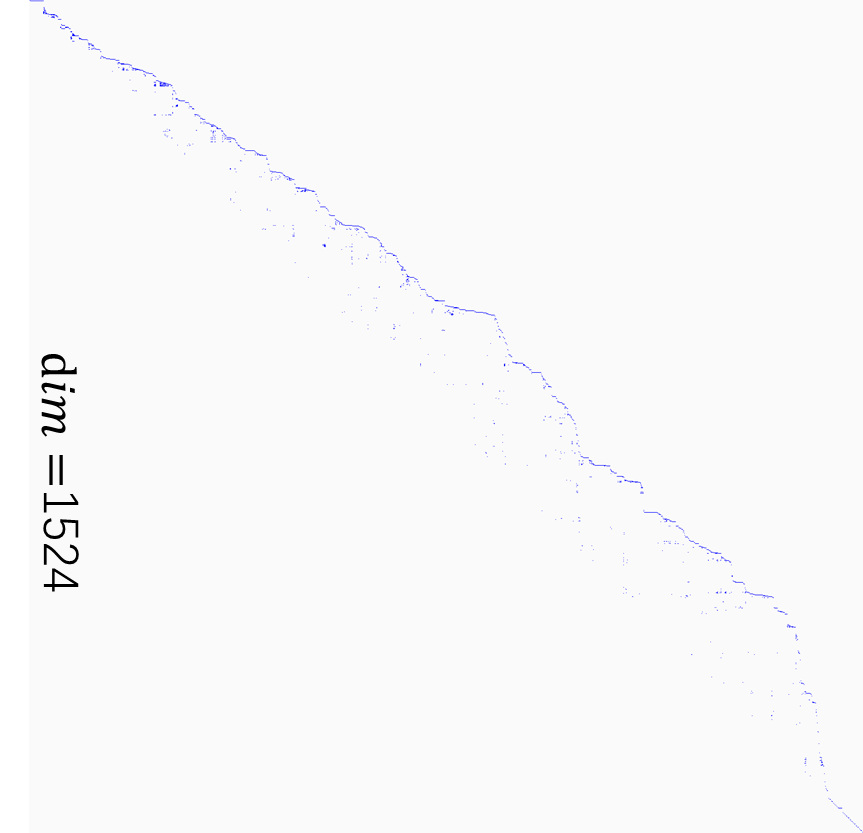} &
            \includegraphics[width=0.3\linewidth]{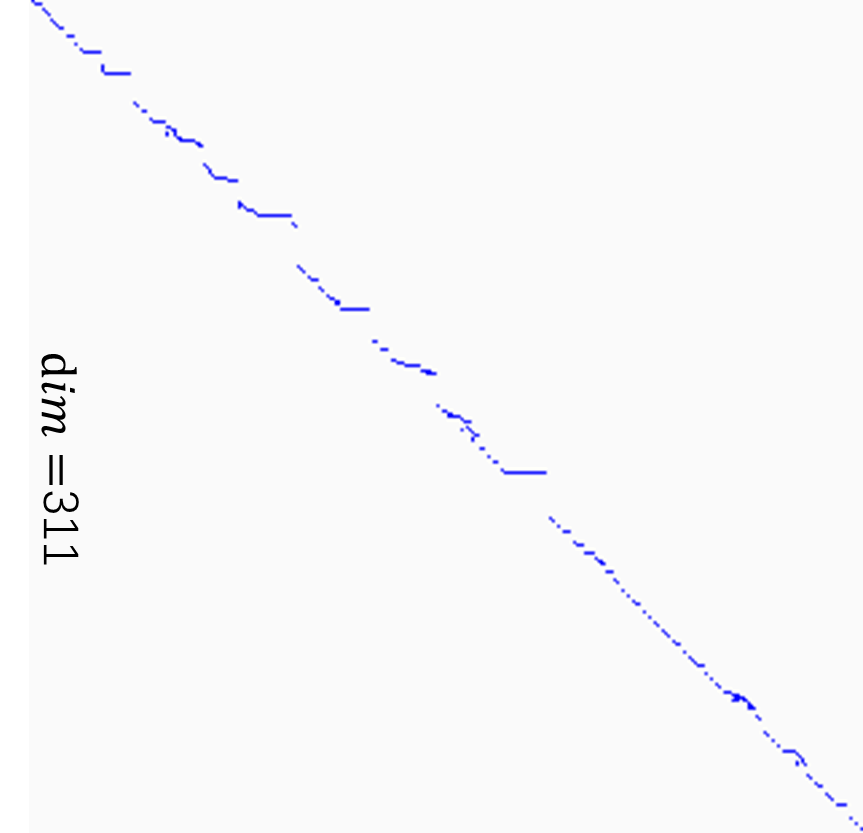}\\
            (d) & (e) & (f)\\
        \end{tabular}
    \end{minipage}
    
    \caption{The adjacent matrix and MBR matrix in schedule block generation for dataset 1: (a) adjacent matrix at initialization; (b) MBR matrix at iteration 1; (c) MBR matrix at iteration 2; (d) MBR matrix at iteration 3; (e) MBR matrix at iteration 4; (f) MBR matrix at iteration 5. The term dim at the left border of each sub-figure indicates the row and column size of the matrix.}
    \label{fig10}
 \end{figure}

\subsubsection{Schedule block generation based on MBR}
\label{sec4.3.2}

View graphs can then be constructed using the retrieved match pairs and used to achieve the schedule block generation and guide cascade hashing feature matching. In the proposed schedule block generation algorithm, the view graph is indicated as the adjacent matrix created from match pairs and MBR matrix by executing matrix band reduction. The former creates the initial connection relationship of original images; the latter represents the compressed structure of permuted images, which is used to facilitate schedule block generation. Figure \ref{fig10} shows the adjacent matrix and MBR matrix during cascade hashing feature matching, in which blue dots indicate corresponding connected images. As shown in Figure \ref{fig10}(a), retrieved match pairs have very sparse connections, as represented by spare blue dots. After applying matrix band reduction at iterations 1, 2, 3, 4, and 5, the MBR matrix shows very compact connections that are almost near the matrix diagonal, as shown in Figure \ref{fig10}(b) to Figure \ref{fig10}(f). Noticeably, during the proceeding of executing feature matching, the dimension of the MBR matrix also decreases from 3743 at iteration 1 to 311 at iteration 5. The main reason is that after feature matching, the images without any other overlapped images have been removed from the MBR matrix. Finally, the dimension of the MBR matrix becomes zero since all match pairs are processed. Therefore, it verifies the correction of the proposed data schedule strategy.

To verify the effectiveness of the proposed data schedule strategy, this study compares feature matching efficiency without and with the data schedule. The only difference between these two tests is the usage of the MBR-based data schedule and SAO-based outlier removal. The performance is evaluated by the metrics \textit{time} and \textit{speed}, which are quantified respectively by time costs of feature matching and the number of pairs processed per second. Table \ref{tab4} presents the results. It is shown that when not using the data schedule strategy, the matching speed is merely 18.94 pairs/s, 23.01 pairs/s, and 23.26 pairs/s, respectively; on the contrary, by using the data schedule strategy, the matching speed improves obviously, which reaches to 340.85 pairs/s, 362.11 pairs/s, and 228.95 pairs/s for each dataset. In other words, the speedup ratio is 18.0, 15.7, and 9.8 for the three datasets, respectively. It can be explained by the low GPU usage for sequential feature matching. Thus, the proposed data schedule strategy improves the efficiency of cascade hashing feature matching.

\begin{table}[!tp]
\centering
\caption{The comparison of feature matching efficiency without and with data schedule. The unit of speed is pairs/s.}
\makebox[\textwidth]{
\begin{tabular}{lrrrrr}
\toprule
\multirow{2}{*}{\textbf{Dataset}} & \multicolumn{2}{c}{\textbf{No schedule}} & \multicolumn{2}{c}{\textbf{Schedule}} & \multirow{2}{*}{\textbf{Speedup}} \\
\cline{2-5}
 & \textbf{Time (s)} & \textbf{Speed} & \textbf{Time (s)} & \textbf{Speed} & \\
\midrule
1 & 3260.22 & 18.9 & 181.20 & 340.8 & 18.0 \\
2 & 2946.06 & 23.0 & 187.20 & 362.1 & 15.7 \\
3 & 15766.86 & 23.3 & 1602.06 & 229.0 & 9.8 \\
\bottomrule
\end{tabular}
}
\label{tab4}
\end{table}

\subsubsection{Ablation study}
\label{sec4.3.3}

The ablation study is conducted for the further evaluation of each key step. In this test, TLBDS is used as the benchmark algorithm because it has almost the same key steps as the proposed solution for feature matching, including data schedule and cascade hashing. Combined with MBR based data schedule and SAO based outlier removal, two versions, i.e., Ours-MBR and Ours-Full, are evaluated. The statistical results are listed in Table \ref{tab5}. It is clearly shown that by using MBR, the feature matching speed of Ours-MBR increases to 235.3 pairs/s, 213.6 pairs/s, and 70.4 pairs/s compared with TLBDS, whose speedup is 4.8, 3.2, and 1.2, respectively, for the three datasets. By using both MBR and SAO, Ours-Full can achieve further acceleration in feature matching, whose speedup is 6.9, 5.5, and 4.0, respectively. We can see that the performance of proposed algorithm can benefit obviously from the usage of these two strategies in feature matching.

\begin{table}[!t]
\centering
\caption{The ablation study of the proposed workflow. The term D1, D2 and D3 represents dataset 1, dataset 2 and dataset 3, respectively.}
\makebox[\textwidth]{
\begin{tabular}{lllllllll}
\toprule
\multirow{2}{*}{\textbf{Method}} & \multicolumn{2}{l}{\textbf{Key step}} & \multicolumn{3}{l}{\textbf{Speed (pairs/s)}} & \multicolumn{3}{l}{\textbf{Speedup}} \\
\cline{2-9}
 & \textbf{MBR} & \textbf{SAO} & \textbf{D1} & \textbf{D2} & \textbf{D3} & \textbf{D1} & \textbf{D2} & \textbf{D3} \\
 \midrule
TLBDS & $\times$ & $\times$ & 49.4 & 66.4 & 56.7 & - & - & - \\
Ours-MBR & \checkmark & $\times$ & 235.3 & 213.6 & 70.4 & 4.8 & 3.2 & 1.2 \\
Ours-Full & \checkmark & \checkmark & 340.8 & 362.1 & 229.0 & 6.9 & 5.5 & 4.0 \\
\bottomrule
\end{tabular}
}
\label{tab5}
\end{table}

\subsection{Feature matching comparison with state-of-the-art methods}
\label{sec4.4}

In this section, the performance of the proposed algorithm is compared with state-of-the-art methods in feaure matching. This test provides the insight comparison of the proposed solution and their fearure matching results.

\subsubsection{The evaluation of feature matching efficiency}
\label{sec4.4.1}

Feature matching efficiency is first evaluated by using two metrics, i.e., \textit{time cost} and \textit{matching speed}. The metric \textit{time cost} includes the time consumption in feature matching and outlier removal, and the metric \textit{matching speed} is calculated as the ratio between match pairs and time costs. In this test, seven algorithms are compared, i.e., ColMap-CPU, ColMap-GPU, AliceVision, TLBDS, GraphPart, Metashape, and Pix4Dmapper. For unbiased comparison, the maximum number of matches is configured as 8,192 in Metashape and Pix4Dmapper as that used in the other algorithms.

\begin{table*}[!t]
    \centering
    \caption{The statistical results of matching efficiency for the evaluated methods in terms of time cost and matching speed.}
    \makebox[\textwidth]{
    \begin{tabular}{lrrrrrr}
        \toprule
        \multirow{2}{*}{\textbf{Method}} & \multicolumn{3}{c}{\textbf{Time cost (min)}} & \multicolumn{3}{c}{\textbf{Matching speed (pairs/s)}} \\
        \cline{2-7}
         & \textbf{Dataset 1} & \textbf{Dataset 2} & \textbf{Dataset 3} & \textbf{Dataset 1} & \textbf{Dataset 2} & \textbf{Dataset 3} \\
        \midrule
        ColMap-CPU & 258.4 & 309.8 & 2064.9 & 4.0 & 3.6 & 3.0 \\
        ColMap-GPU & 3.4 & 3.6 & 71.7 & 302.4 & 315.9 & 85.2 \\
        AliceVision & 22.7 & 24.6 & 155.3 & 61.3 & 58.5 & 55.8 \\
        TLBDS & 11.3 & 12.1 & 70.8 & 91.1 & 93.7 & 86.4 \\
        GraphPart & 8.3 & 10.4 & 54.6 & 124.0 & 108.6 & 112.0 \\
        Metashape & 8.3 & 9.2 & 57.0 & 33.0 & 38.9 & 20.3 \\
        Pix4Dmapper & 21.0 & 10.9 & 136.7 & 41.0 & 54.6 & 30.2 \\
        Ours & 3.0 & 3.1 & 26.7 & 340.8 & 362.1 & 229.0 \\
        \bottomrule
    \end{tabular}
    }
    \label{tab6}
\end{table*}

\begin{figure}[!t]
    \centering
    \includegraphics[width=1.0\linewidth]{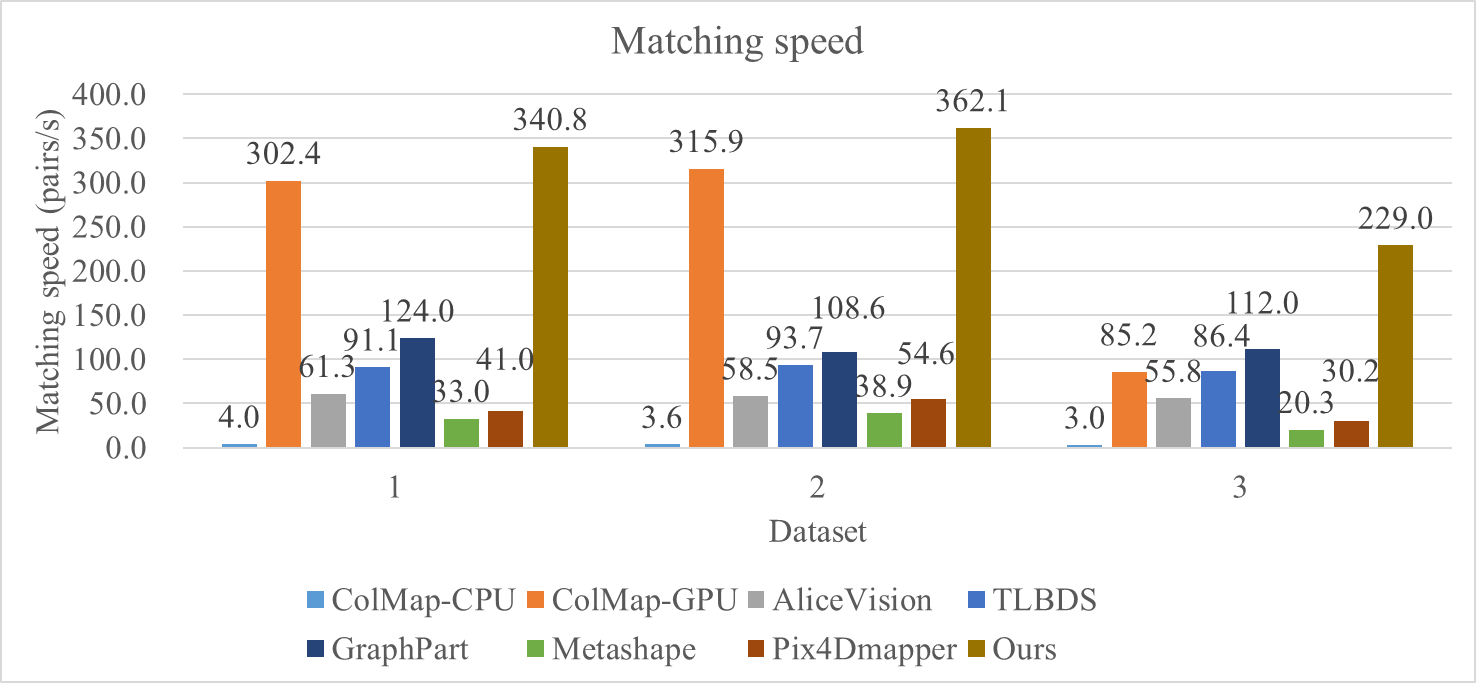}
    \caption{The feature matching speed of all evaluated methods.}
    \label{fig11}
 \end{figure}

 Table \ref{tab6} presents the statistical results of time cost and matching speed of all the evaluated algorithms, and Figure \ref{fig11} plots matching speed for a visual comparison. Considering the metric time cost, ColMap-CPU has the lowest efficiency, which is 258.4 mins, 309.8 mins, and 2064.9 mins for the three datasets, respectively. The main reason is the usage of multi-core CPU for ANN-based feature matching. When accelerated by CPU-based cascade hashing, the time cost of AliceVision obviously decreases to 22.7 mins, 24.6 mins, and 155.3 mins with a speedup ratio of 11.4, 12.6, and 13.3 for the three datasets, respectively. Furthermore, by using the GPU-based cascade hashing, the efficiency of TLBDS is then accelerated by approximately two times when compared with that of AliceVision. Therefore, the test results verify the effectiveness of cascade hashing for feature matching.

\begin{figure}[!t]
    \centering
    \begin{minipage}[t]{1.0\linewidth}
    \centering
        \begin{tabular}{@{\extracolsep{\fill}}c@{}c@{}c@{}@{\extracolsep{\fill}}}
            \includegraphics[width=0.45\linewidth]{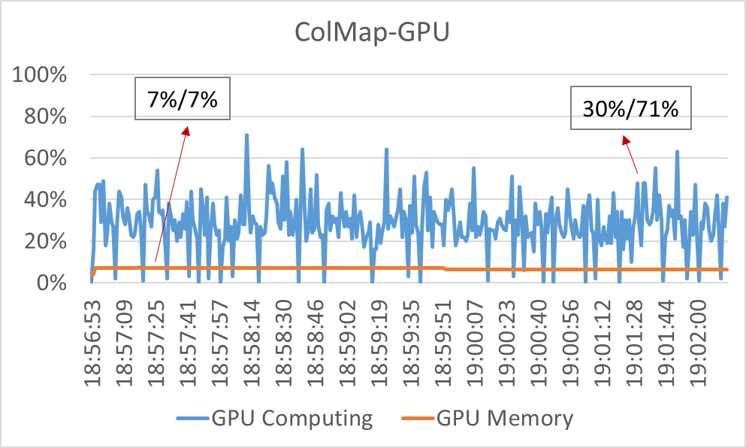} &
            \includegraphics[width=0.45\linewidth]{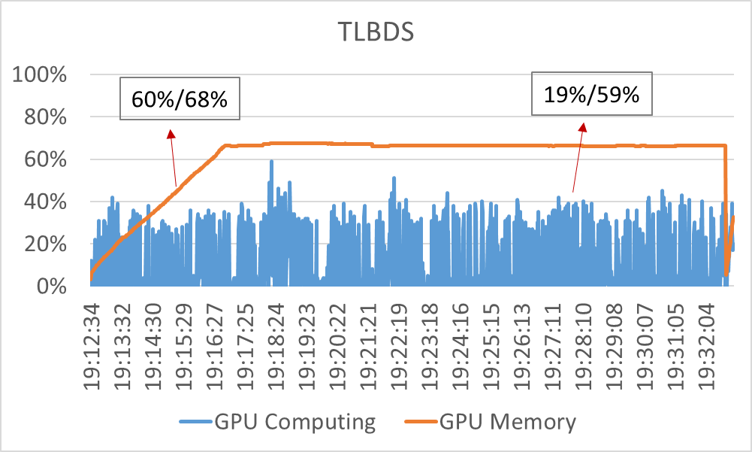} \\
            (a) & (b)\\
        \end{tabular}
    \end{minipage}
    \begin{minipage}[t]{1.0\linewidth}
    \centering
        \begin{tabular}{@{\extracolsep{\fill}}c@{}c@{}c@{}@{\extracolsep{\fill}}}
            \includegraphics[width=0.45\linewidth]{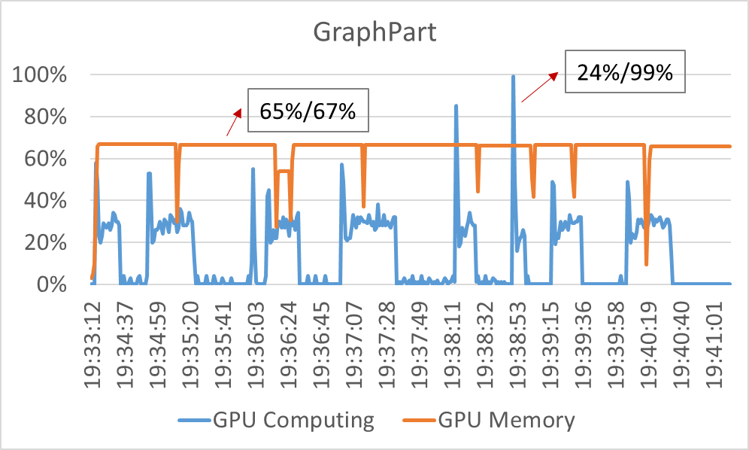} & 
            \includegraphics[width=0.45\linewidth]{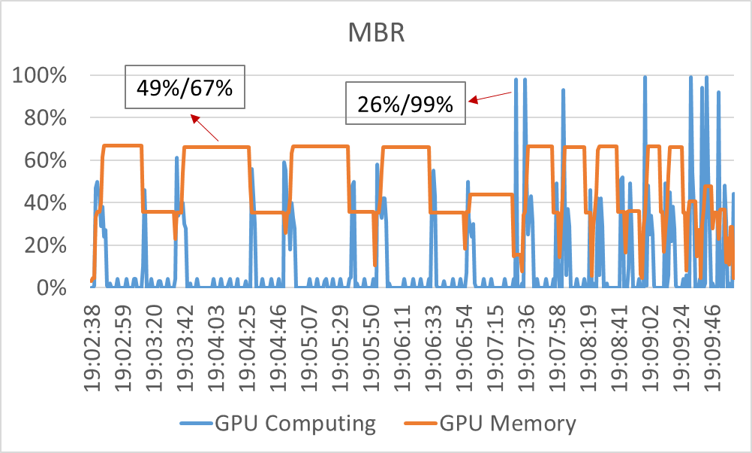} \\
            (c) & (d)\\
        \end{tabular}
    \end{minipage}
    \caption{The comparison of the GPU computing and memory usage of dataset 1: (a) ColMap-GPU; (b) TLDBS; (c) GraphPart; (d) MBR. The values in the rectangle are the statitics of mean and max, respectively, for each evaluated method.}
    \label{fig11a}
 \end{figure}

For the further analysis, we can see that: (1) the time cost of ColMap-GPU is approximately one third of that in TLBDS for datasets 1 and 2, while they have almost the same time cost in dataset 3. The main reason is due to their different data schedule strategies. In TLBDS, images are ordered according to their connections and are sequentially fed into GPU during feature matching. Although it has a higher usage ratio of GPU memory, the weak connection between images causes a lower usage ratio of GPU computing, as shown in Figure \mbox{\ref{fig11a}(b)}. On the contrary, ColMap-GPU divides match pairs into blocks that are fed into GPU at one time, which increases overall usage ratio of GPU computing. However, as the data schedule strategy in ColMap-GPU does not consider image connections, the low usage ratio of GPU memory can be observed in Figure \mbox{\ref{fig11a}(a)}; (2) compared with ColMap-GPU, TLBDS, and GraphPart, the proposed algorithm achieves the highest efficiency for the three datasets, whose time cost is 3.0 mins, 3.1 mins, and 26.7 mins, respectively. The main reason can be explained by the MBR-based data schedule, which considers the usage ratio of GPU computing and memory to increase feature matching efficiency, as presented in Figure \mbox{\ref{fig11a}(d)}.

  \begin{figure}[!t]
    \centering
    \begin{minipage}[t]{1.0\linewidth}
    \centering
        \begin{tabular}{@{\extracolsep{\fill}}c@{}c@{}c@{}@{\extracolsep{\fill}}}
            \includegraphics[width=0.45\linewidth]{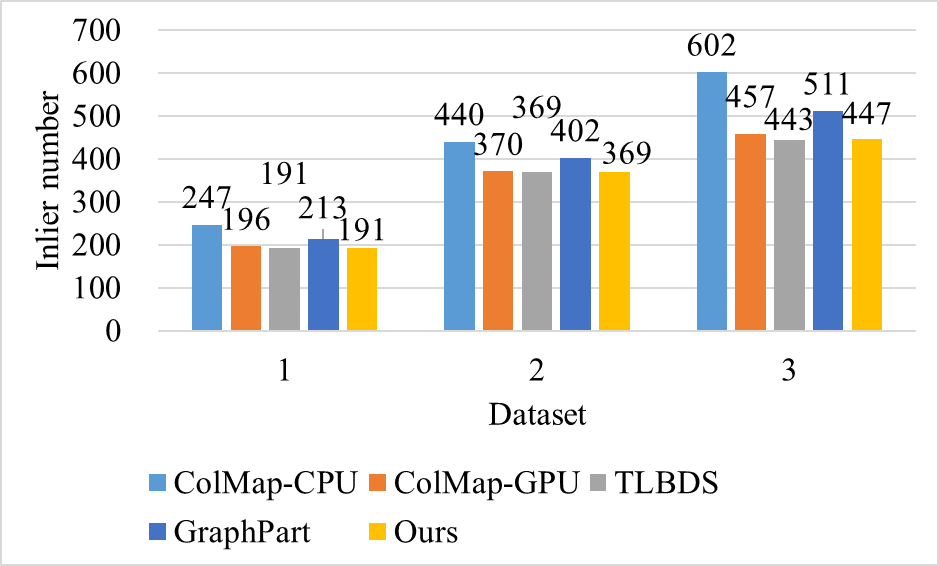} &
            \includegraphics[width=0.45\linewidth]{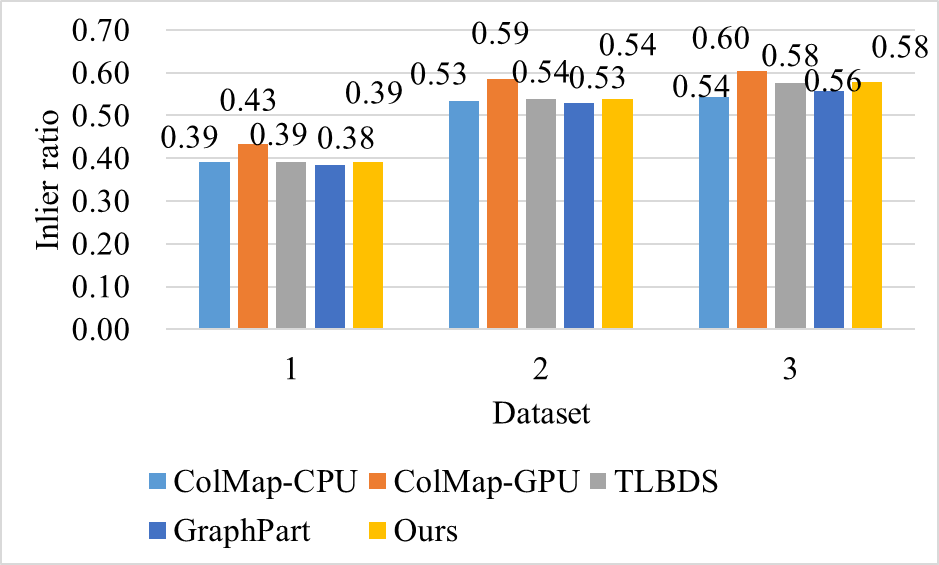}\\
            (a) & (b)\\
        \end{tabular}
    \end{minipage}
   
    \caption{The comparison of inlier number and inlier ratio for the evaluated algorithms: (a) inlier number; (b) inlier ratio.}
    \label{fig12}
 \end{figure}

  \begin{figure}[!t]
    \centering
    \begin{minipage}[t]{1.0\linewidth}
    \centering
        \begin{tabular}{@{\extracolsep{\fill}}c@{}c@{}c@{}@{\extracolsep{\fill}}}
            Dataset 1 & Dataset 2 & Dataset 3\\
            \includegraphics[width=0.3\linewidth]{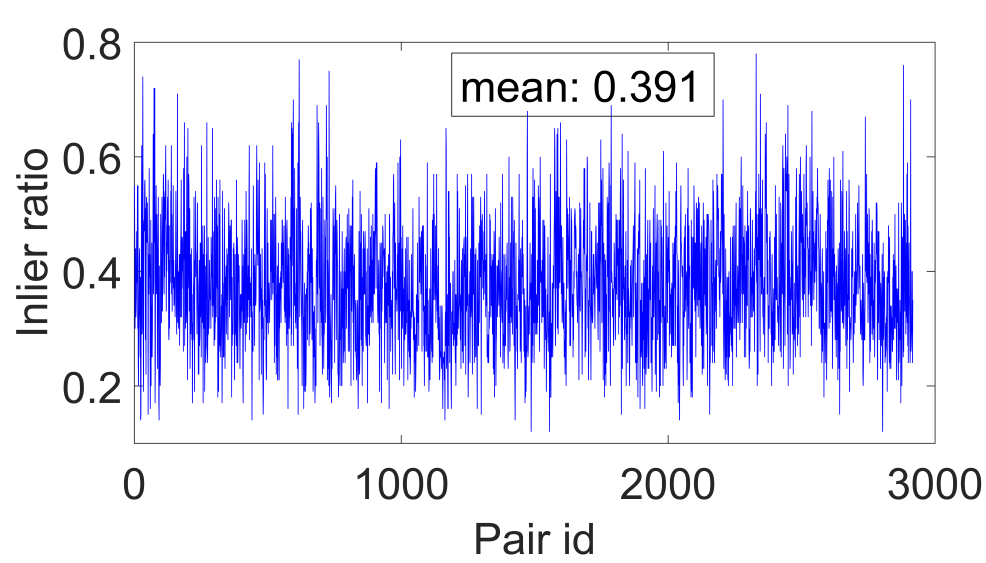} &
            \includegraphics[width=0.3\linewidth]{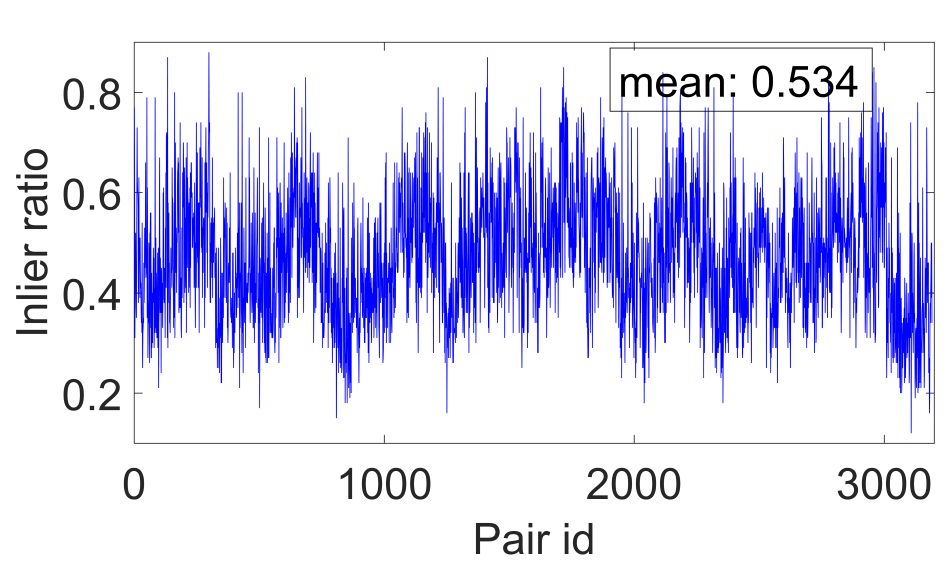} &
            \includegraphics[width=0.3\linewidth]{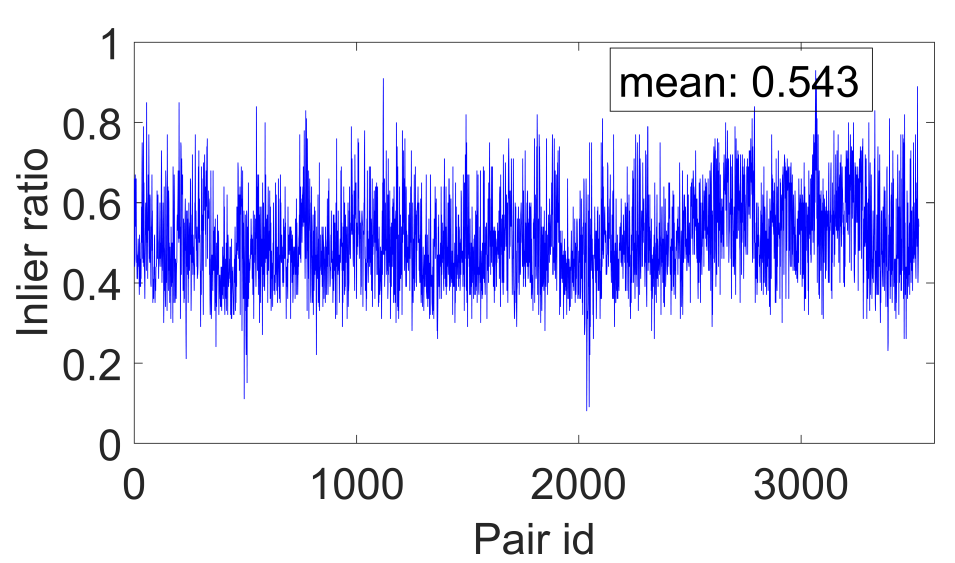}\\
            (a) & (b) & (c)\\
        \end{tabular}
    \end{minipage}

    \begin{minipage}[t]{1.0\linewidth}
    \centering
        \begin{tabular}{@{\extracolsep{\fill}}c@{}c@{}c@{}@{\extracolsep{\fill}}}
            \includegraphics[width=0.3\linewidth]{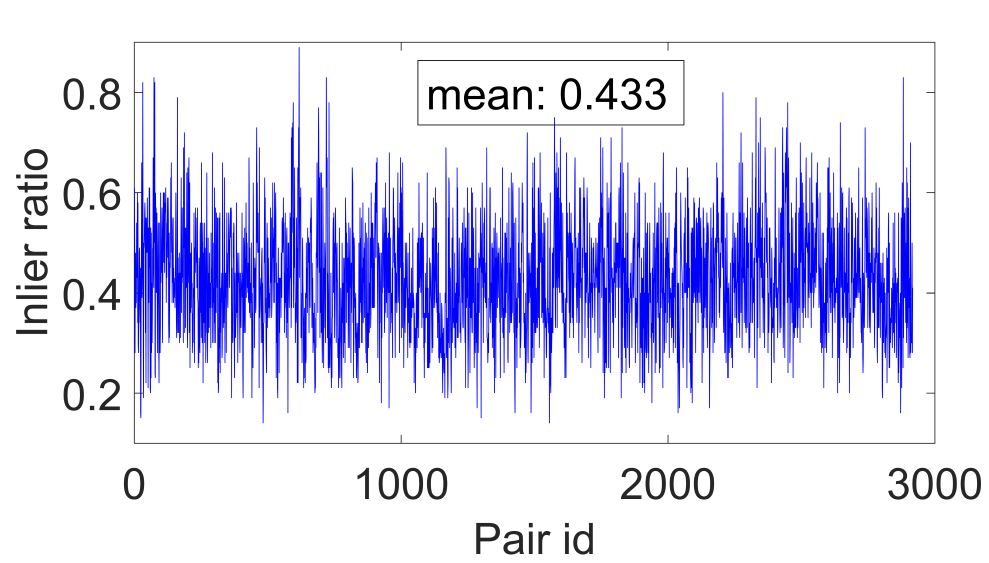} &
            \includegraphics[width=0.3\linewidth]{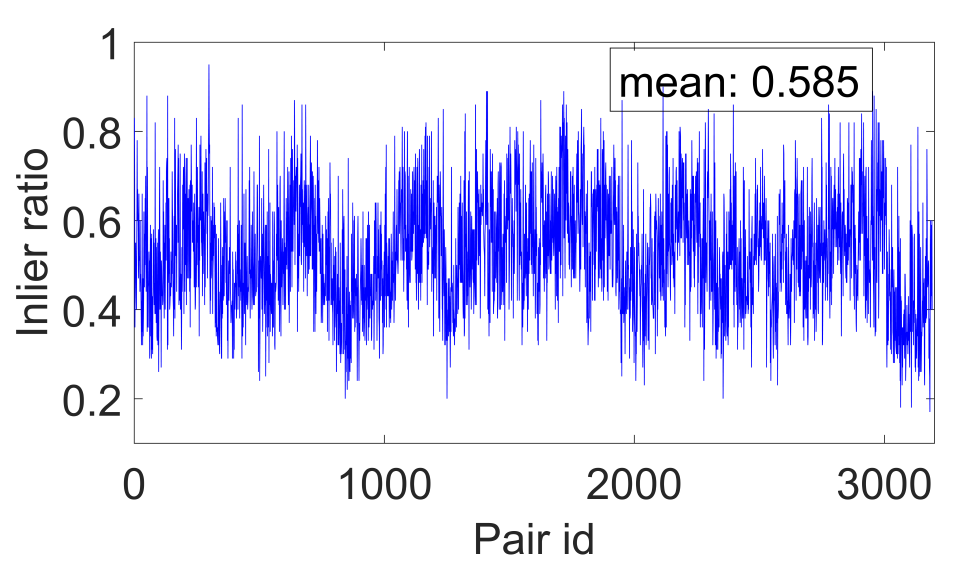} &
            \includegraphics[width=0.3\linewidth]{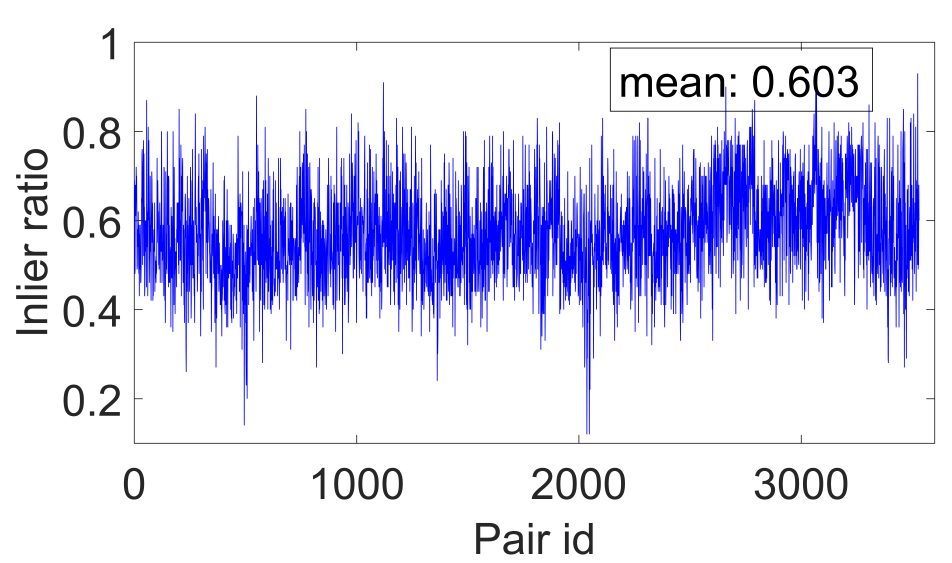}\\
            (d) & (e) & (f)\\
        \end{tabular}
    \end{minipage}

    \begin{minipage}[t]{1.0\linewidth}
    \centering
        \begin{tabular}{@{\extracolsep{\fill}}c@{}c@{}c@{}@{\extracolsep{\fill}}}
            \includegraphics[width=0.3\linewidth]{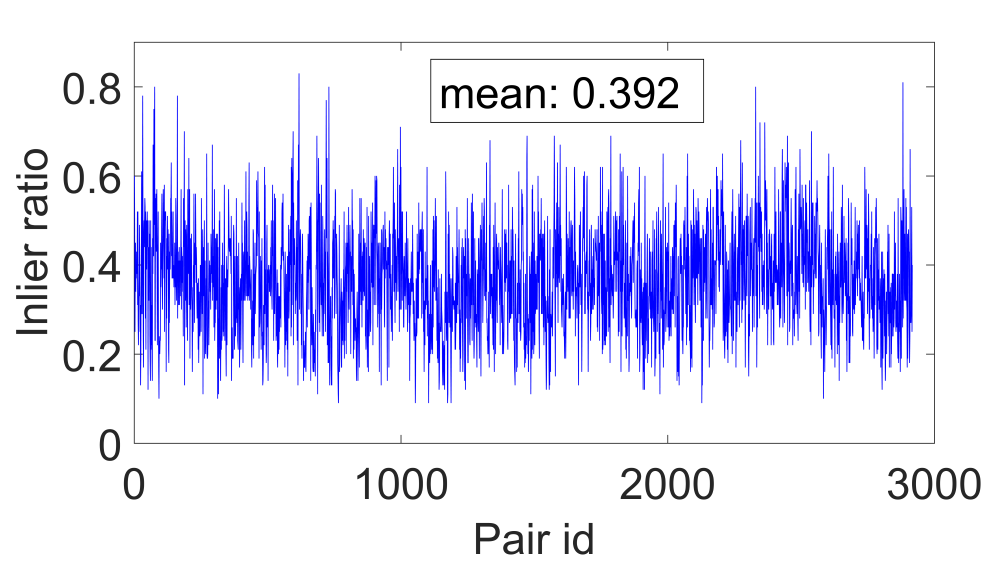} &
            \includegraphics[width=0.3\linewidth]{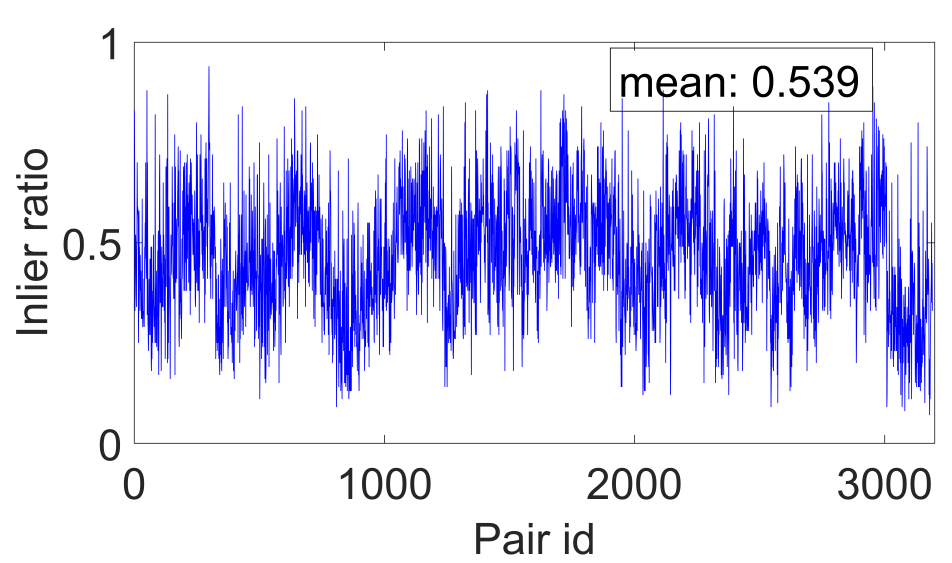} &
            \includegraphics[width=0.3\linewidth]{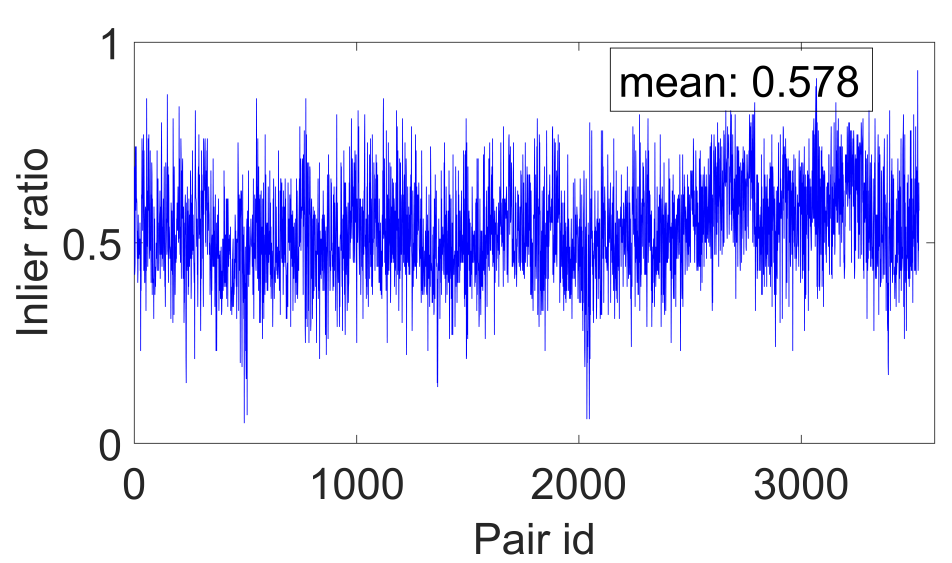}\\
            (g) & (h) & (i)\\
        \end{tabular}
    \end{minipage}
    
    \caption{The individual inlier ratio for the three datasets: (a), (b) and (c) for ColMap-CPU; (d), (e) and (f) for ColMap-GPU; (g), (h) and (i) for the proposed algorithm.}
    \label{fig13}
 \end{figure}

\subsubsection{The evaluation of feature matching resutls}
\label{sec4.4.2}

Feature matching results are then evaluated by using two metrics, i.e., \textit{inlier number} and \textit{inlier ratio}, which represents the average true matches and the average number ratio between inliers and initial matches over all match pairs, respectively. Figure \ref{fig12} presents the comparison. Noticeably, these two metrics are not calculated for AliceVision, Metashape, and Pix4Dmapper since only the executable application is utilized in this test. It is shown that for the metric \textit{inlier number}, ColMap-GPU, TLBDS, and the proposed algorithm achieve comparable performance, which is relatively lower than that of ColMap-CPU. The main reason can be the usage of GPU with the single-precision floating-point format for feature matching when compared with CPU with the double-precision floating-point format. For the metric \textit{inlier ratio}, ColMap-GPU achieves the highest performance, which is 0.43, 0.59, and 0.60 for the three datasets. From the setting in Table \mbox{\ref{tab2}}, it can be explained by the used feature matching algorithm. In contrast to ColMap-CPU, TLBDS and the proposed algorithm with ANN for feature matching, ColMap-GPU uses NNS for feature matching, which has higher precision generally. In addition, Figure \ref{fig13} presents the individual plot of the inlier ratio for the three datasets. Since TLBDS uses the same cascade hashing for feature matching, only the result of the proposed algorithm is analyzed. We can see that the proposed algorithm achieves better performance compared with ColMap-CPU, and evaluated methods have almost consistent change in inlier ratios for all match pairs. In conclusion, the proposed algorithm achieves speedup ratios ranging from 77.0 to 100.0 for feature matching compared with traditional KD-Tree based matching methods and provides comparable matching results for subsequent SfM reconstruction.

\subsection{SfM-based 3D reconstruction}
\label{sec4.5}

SfM-based 3D reconstruction is then executed using feature matching results. In general, the performance of SfM-based 3D reconstruction is affected by the inlier number and correspondence distribution over image planes, which can be utilized to evaluate the quality of feature matching. SfM-based 3D reconstruction can be executed without GCPs for relative orientation and with GCP for absolute orientation. The details are presented as follows.

\subsubsection{Relative BA without GCPs}
\label{sec4.5.1}

\begin{table}[!t]
	\centering
	\caption{The statistical results of relative BA without GCPs in terms of time cost, the number of registered images, and the number of reconstructed 3D points.}
	\makebox[\linewidth]{
		\begin{tabular}{llrrr}
			\toprule
			\textbf{Metric} & \textbf{Method} & \textbf{Dataset 1} & \textbf{Dataset2} & \textbf{Dataset 3} \\
			\midrule
                \multirow{5}{*}{\begin{tabular}[c]{@{}l@{}}Time cost\\ (min)\end{tabular}}
                &ColMap-GPU&	13.9&	46.1&	703.2\\
                &TLBDS&	14.8&	44.6&	714.3\\
                &Metashape&	41.8& 	69.9 &	—\\
                &Pix4Dmapper&	234.2&	352.4&	—\\
                &Ours&	13.9&	52.4&	669.5\\

			\midrule
			\multirow{5}{*}{\begin{tabular}[c]{@{}l@{}}\#Registered\\ images\end{tabular}  } 
                &ColMap-GPU&	3,736&	4,029&	21,624\\
                &TLBDS&	3,734&	4,027&	21,349\\
                &Metashape&	3,743&	4,029&	—\\
                &Pix4Dmapper&	3736&	4027&	—\\
                &Ours&	3,728&	4,027&	21,627\\

			\midrule
			\multirow{5}{*}{\begin{tabular}[c]{@{}l@{}}\#3D points\end{tabular}} 
                &ColMap-GPU&	874,869&	1,340,548&	7,698,043\\
                &TLBDS&	898,650&	1,394,779&	8,063,822\\
                &Metashape&	2,824,443&	2,532,249&	—\\
                &Pix4Dmapper&	4,407,343&	5,206,527&	—\\
                &Ours&	896,669&	1,403,623&	8,100,207\\
			\bottomrule
		\end{tabular}
		\label{tab7}
  }
\end{table}

\begin{figure}[!t]
    \centering
    
    \begin{minipage}[t]{1.0\linewidth}
    \centering
            \includegraphics[width=1.0\linewidth]{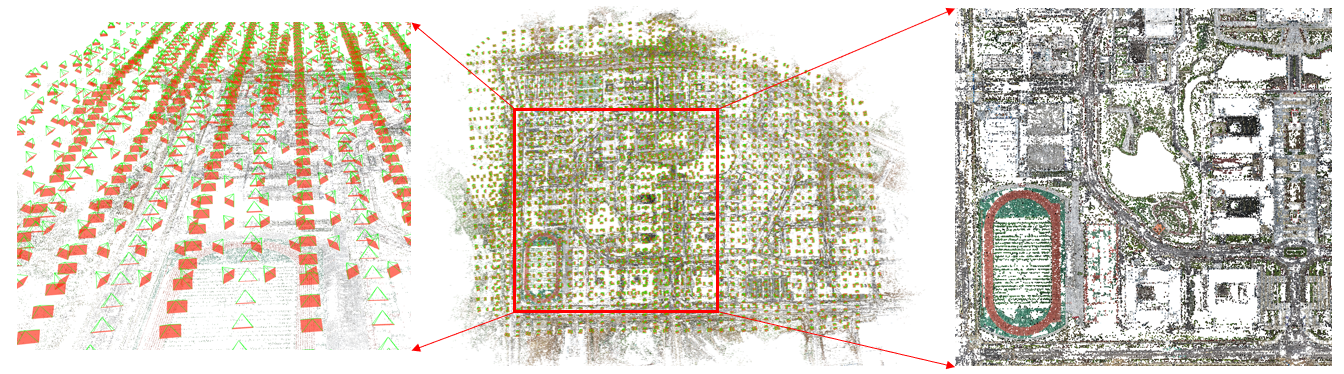}\\
            (a)\\
    \end{minipage}

    \begin{minipage}[t]{1.0\linewidth}
    \centering
            \includegraphics[width=1.0\linewidth]{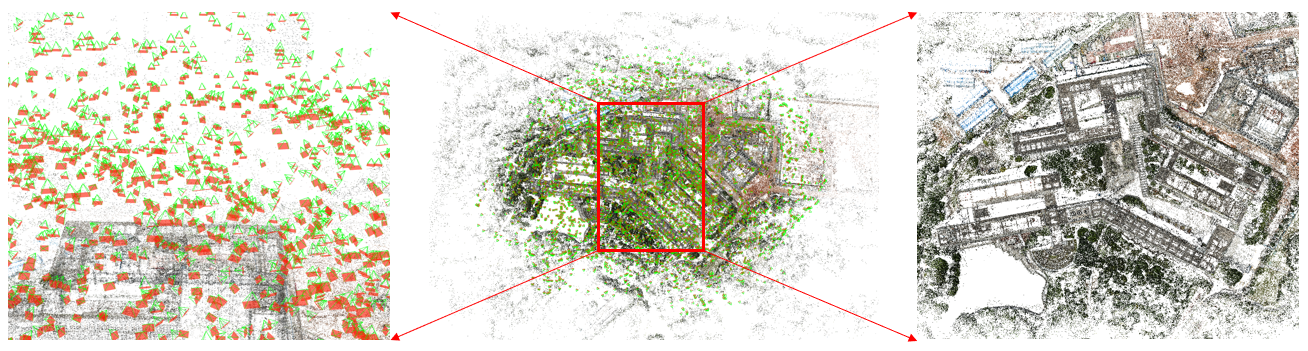}\\
            (b)\\
    \end{minipage}

    \begin{minipage}[t]{1.0\linewidth}
    \centering
            \includegraphics[width=1.0\linewidth]{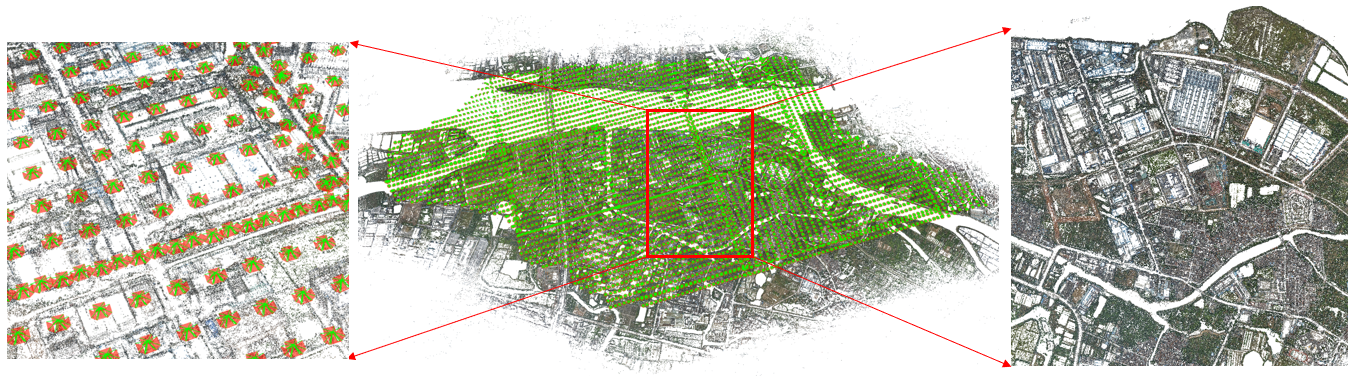}\\
            (c)\\
    \end{minipage}
    
    \caption{SfM reconstruction results of the three datasets: (a) dataset 1; (b) dataset 2; (c) dataset 3. The sub-figures from the left to right side represents oriented camera poses, entire SfM reconstruction, and detailed 3D points, respectively.}
    \label{fig14}
 \end{figure}

For relative BA without GCPs, SfM is used to compare the efficiency and completeness of reconstructed models, in which the efficiency is quantified by time costs, and the completeness is measured by the number of registered images and 3D points. In this test, the parallel SfM solution proposed in \cite{jiang2024efficient} is used as the engine for ColMap-GPU, TLBDS, and the proposed algorithm. 

Table \ref{tab7} presents the statistical results of relative BA without GCPs for the three datasets. It is shown that except for the two commercial software packages Metashape and Pix4Dmapper, all the other algorithms have comparable time costs consumed in the three datasets, which are 13.9 mins, 52.4 mins, and 669.5 mins for the proposed algorithm. The same findings can also be observed from the metric of the number of registered images, except for TLBDS in dataset 3. For the metric of the number of 3D points, more 3D points are reconstructed in Metashape and Pix4Dmapper since they use different feature detection and matching strategies. For the further comparison between ColMap-GPU and the proposed algorithm, we can see that more 3D points are resumed in the proposed algorithm although almost the same number of images are registered in these two algorithms. The main reason can be explained by the guided feature matching in cascade hashing that improves correspondence distribution. For visual inspection, the reconstructed 3D models are presented in Figure \ref{fig14}. In conclusion, the proposed algorithm can provide reliable and enough feature matches for SfM reconstruction.

\subsubsection{Absolute BA with GCPs}
\label{sec4.5.2}

\begin{table*}[t!]
	\centering
	\caption{The statistical results of absolute BA with GCPs for dataset 2.}
    \makebox[\linewidth]{
	\begin{tabular}{llllllllll}
		\toprule
		\multirow{2}{*}{\textbf{Method}} & \multicolumn{3}{c}{\textbf{Max (m)}} & \multicolumn{3}{c}{\textbf{Mean (m)}} & \multicolumn{3}{c}{\textbf{Std.dev. (m)}} \\
		\cline{2-10}
		& \textbf{$|X|$} & \textbf{$|Y|$} & \textbf{$|Z|$} & \textbf{$|X|$} & \textbf{$|Y|$} & \textbf{$|Z|$} & \textbf{$|X|$} & \textbf{$|Y|$} & \textbf{$|Z|$} \\
		\midrule 
            ColMap-GPU&	0.129&	0.101 &	0.196 &	0.054& 	0.050& 	0.085 &	0.063& 	0.058& 	0.063 \\
            TLBDS&	0.150&	0.151 &	0.193&	0.034& 	0.050 &	0.085 &	0.053 &	0.067 &	0.059 \\
            Pix4dMapper&	0.028&	0.036&	0.048&	0.010 &	0.012&	0.015 &	0.013 &	0.016 &	0.019 \\
            Ours&	0.065&	0.123 &	0.096 &	0.018& 	0.026& 	0.037 &	0.024 &	0.040& 	0.037 \\
		\bottomrule
	\end{tabular}
	\label{tab8}
 }
\end{table*}

\begin{figure}[t!]
    \centering
    
    \begin{minipage}[t]{1.0\linewidth}
    \centering
            \includegraphics[width=0.7\linewidth]{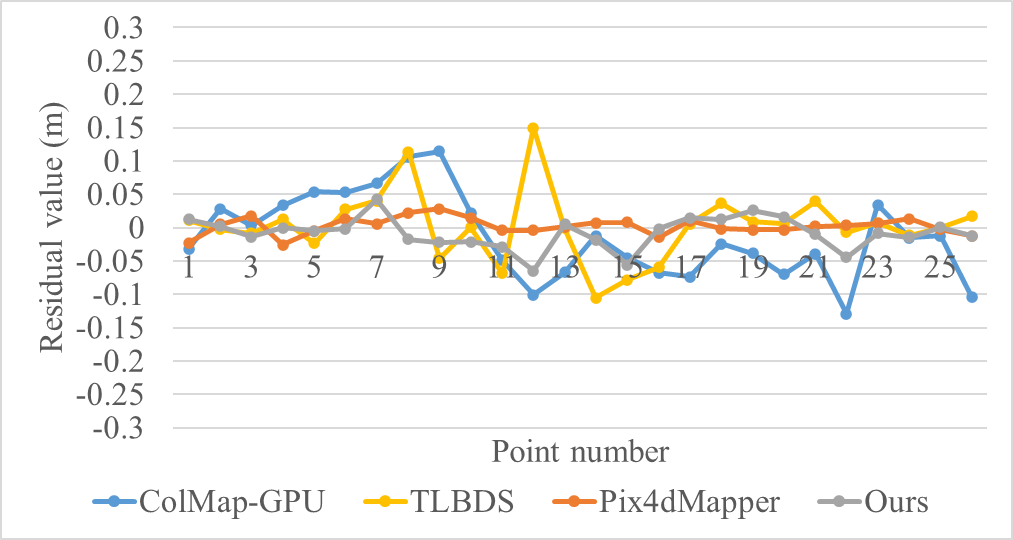}\\
            (a)\\
    \end{minipage}

    \begin{minipage}[t]{1.0\linewidth}
    \centering
            \includegraphics[width=0.7\linewidth]{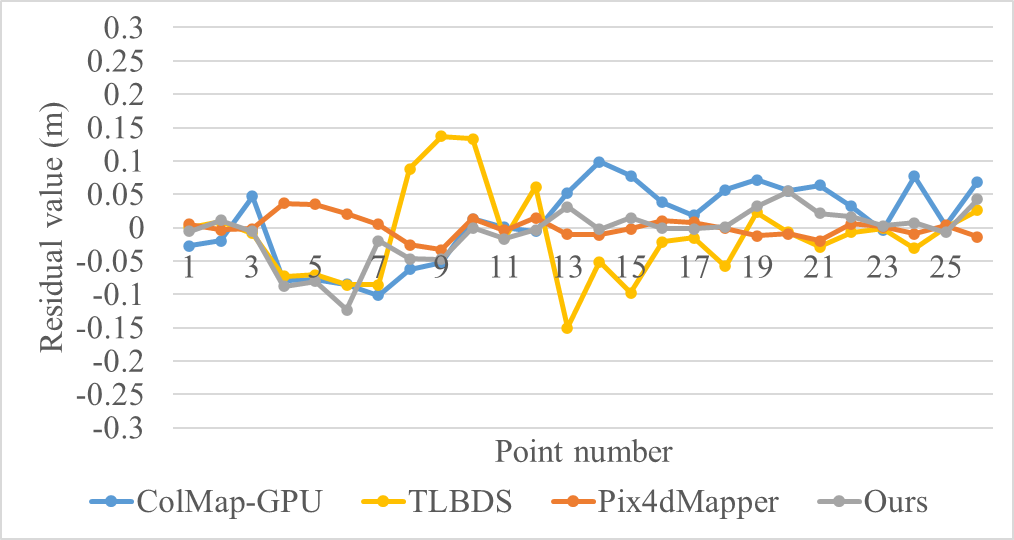}\\
            (b)\\
    \end{minipage}

    \begin{minipage}[t]{1.0\linewidth}
    \centering
            \includegraphics[width=0.7\linewidth]{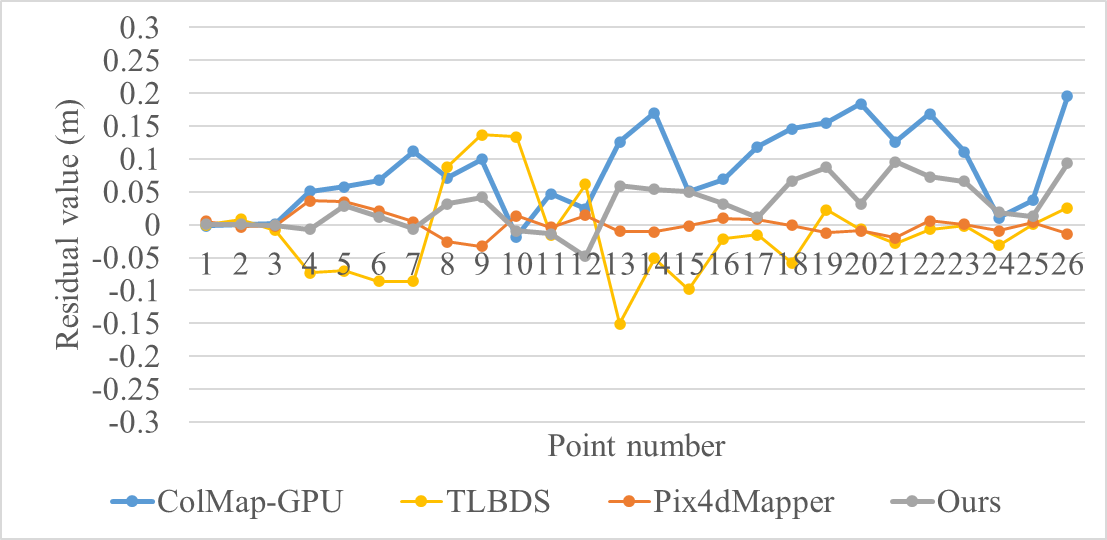}\\
            (c)\\
    \end{minipage}
    
    \caption{The residual plots after absolute BA by using GCPs in dataset 2: (a) the residual plot in the X axis; (b) the residual plot in the Y axis; (c) the residual plot in the Z axis.}
    \label{fig15}
 \end{figure}

With GCPs aided SfM reconstruction, absolute orientation can be conducted to evaluate model geo-referencing accuracy. In this test, three evenly distributed GCPs among the 26 GCPs surveyed in dataset 2 are used as control points for model geo-referencing, and the others are used as check points for accuracy evaluation. The residual is calculated as the positioning offset between estimated model points and corresponding check points. Besides, three metrics, i.e., \textit{Max}, \textit{Mean}, and \textit{Std.dev}, are used for performance evaluation, which indicates the maximum, average, and standard deviation of residuals in the X, Y, and Z directions, respectively.

Table \ref{tab8} presents the statistical results of absolute BA with GCPs, and Figure \ref{fig15} shows the residual plots. We can conclude that: (1) Pix4Dmapper achieves the highest precision among all evaluated algorithms, whose \textit{Std.dev} respectively reaches 0.013 m, 0.016 m, and 0.019 m in the X, Y, and Z directions; (2) TLBDS and ColMap-GPU have comparable precision in the Y and Z directions, while TLBDS achieves higher precision in the X direction; (3) the proposed algorithm ranks second among all evaluated methods, and its precision is 0.024 m, 0.040 m, and 0.037 m in the three directions. It can also be verified by the consistent residual plots shown in Figure \ref{fig15}. Considering that the GSD of dataset 2 is 1.2 cm, the geo-referencing accuracy of the proposed algorithm is better than 2.0 and 3.5 times the GSD value in the horizontal and vertical directions, respectively. Thus, the proposed algorithm can be an efficient and reliable solution for feature matching of large-scale UAV images.

\section{Discussion}
\label{sec5}
\added{In this study, we propose a GPU data schedule algorithm for efficient feature matching of large-scale UAV images. The main issue addressed by the proposed algorithm is to make a balance between redundant data IO burden and the usage of GPU computing power. For this purpose, two major contributions have been made to the proposed solution. On the one hand, by using VLAD-HNSW image retrieval for view graph construction, we propose a MBR based data schedule strategy to divide the sparsely connected view graph into compact sub-graphs. The data schedule strategy adapts well to the volume of GPU memory as well as the connection structure of images; on the other hand, using the data schedule strategy, we further implement a cascade hashing based feature matching workflow, which integrates the SAO-based local constraint and RANSAC-based global verification for efficient and reliable outlier removal. By using large-scale UAV datasets captured by classical oblique and optimized views photogrammetry, the performance of the proposed algorithm has been extensively evaluated in terms of feature matching and SfM-based 3D reconstruction and compared with state-of-the-art open-source and commercial software packages.}

\added{For the proposed algorithm, there are two critical parameters, i.e., the schedule block size $Size_{blk}$ for data schedule and the ratio test threshold $t_r$ for outlier removal. According to the test results presented in Section \ref{sec4.2}, we can see that for UAV datasets with dense connections, the schedule block size influences the overall efficiency of feature matching, and a large value can increase the feature matching efficiency; on the contrary, it almost has no influence on the datasets with sparse connections, as verified by the results in Figure \ref{fig6}. In practice, a majority of UAV datasets have been captured according to the rule the classical oblique photogrammetry. The proposed algorithm can perform well with a proper schedule block size that fits the GPU memory volume. For the ratio test threshold, we can see that the optimal value of 0.5 has been verified by the results presented in Figure \ref{fig6}, which differs from the setting in other libraries, e.g., 0.8 in SIFTGPU. The main reason can be explained by the pre-filtering of cascade hashing mapping. In addition, the experimental results in Section \ref{sec4.4} demonstrate that the proposed solution achieves the highest efficiency in feature matching and provides reliable matching results for SfM-based 3D reconstruction. On the one hand, it can obviously increase the efficiency of feature matching with speedup ratios ranging from 77.0 to 100.0 compared with traditional KD-Tree based matching methods. The major advantages lie in the MBR-based data schedule and GPU-accelerated cascade hashing; on the other hand, the proposed solution can achieve comparable relative BA results in terms of the number of registered images and 3D points and better performance in absolute BA when compared with ColMap-GPU and TLBDS. In conclusion, the proposed algorithm can be an efficient and reliable solution for feature matching of large-scale UAV images.}

\added{For the proposed feature matching algorithm, some issues should be paid attention. First, VLAD-HNSW based image retrieval has been designed for match pair selection, and its performance has been validated using large-scale UAV datasets. The sparsity of image networks has an impact on MBR-based feature matching performance, but not on image retrieval performance. As an extreme case, the linear flight path causes the image adjacency matrix with non-zero values only along the diagonal. In this case, the data schedule problem is simplified, which indicates that the MBR-based data schedule strategy has no obvious advantage over other algorithms. Second, feature matching based on cascade hashing would perform worse in the presence of low texture or repeating patterns. Almost every other feature matching algorithm does the same. The SAO constraint had been designed for this study in order to remove outliers following the initial feature matching.  SAO-based outlier elimination can also be effective, even when low-texture or repeating patterns would reduce the quantity and inlier ratio of initial matches.}

\section{Conclusions}
\label{sec6}

This study proposes matrix band reduction-based GPU data schedule for efficient feaure matching of UAV images. The core idea is to divide the whole dataset into blocks via matrix band reduction and achieve efficient feature matching based on GPU-accelerated cascade hashing. First, by using the VLAD-HNSW image retrieval technique, a view graph is created to establish image connections, which is then divided into compact blocks via matrix band reduction. The data schedule strategy adapts well to the connection structure of images and the volume of GPU memory. Second, guided by the generated blocks, feature matching is executed sequentially based on GPU-accelerated cascade hashing, in which initial candidate matches are refined by the SAO-based local constraint and RANSAC-based global verification in outlier removal. Finally, the proposed solution is evaluated and compared by using large-scale UAV datasets. The results demonstrate that it can achieve efficient feature matching with tens to hundreds of speedup ratios and provide reliable matches for SfM-based 3D reconstruction.

\section*{Acknowledgments}
This research was funded by the National Natural Science Foundation of China (Grant No. 42371442), Shenzhen Science and Technology Program (Grant No. JCYJ20250604181614019), and Shenzhen Key Laboratory Program (Grant No. SYSPG20241211173845013).

\appendix

\section{}

\added{
The procedure of the GPS algorithm is presented in Algorithm \ref{algorithm2}, and the code in C++ is released at \url{https://github.com/json87/gps4mbr}.
}

\begin{algorithm}
		\caption{Gibbs-Poole-Stockmeyer (GPS) algorithm}
        \label{algorithm2}
		\begin{algorithmic}[1]
			\State \textbf{Input:} A connected graph $G = (V, E)$
			\State \textbf{Output:} A permutation $p$ that reduces the matrix bandwidth
			
			\State \textbf{procedure} \textsc{PseudoPeripheralNodeSelection}($G$)
			\State \quad Select an initial node $u$ with minimum degree
			\State \quad Generate the level structure $L(u)$ rooted at $u$
			\State \quad \textbf{while} true \textbf{do}
			\State \quad \quad Select a node $v$ in the last level of $L(u)$ with minimum degree
			\State \quad \quad Generate the level structure $L(v)$ rooted at $v$
			\State \quad \quad \textbf{if} depth($L(v)$) $>$ depth($L(u)$) \textbf{then}
			\State \quad \quad \quad $u \leftarrow v$, $L(u) \leftarrow L(v)$
			\State \quad \quad \textbf{else}
			\State \quad \quad \quad \textbf{return} $(u, v)$ as the pseudo-peripheral pair
			\State \quad \quad \textbf{end if}
			\State \quad \textbf{end while}
			
			\State \textbf{procedure} \textsc{GpsOrdering}($G, u, v$)
			\State \quad Let $L_u$ and $L_v$ be the level structures rooted at $u$ and $v$
			\State \quad Combine $L_u$ and $L_v$ to create a combined level structure $C$
			\State \quad \textbf{for each} level $i$ in $C$ \textbf{do}
			\State \quad \quad Sort nodes in level $i$ based on their degrees and connection to previous levels
			\State \quad \textbf{end for}
			\State \quad Number nodes sequentially from level 1 to $k$ to form permutation $p$
			\State \quad \textbf{return} $p$
		\end{algorithmic}
	\end{algorithm}

\bibliographystyle{elsarticle-harv}
\bibliography{mybibfile}





\end{document}